\documentclass[runningheads]{llncs}

\usepackage{eccv}

\usepackage{eccvabbrv}

\usepackage{graphicx}
\usepackage{booktabs}
\usepackage{enumitem}
\usepackage{wrapfig}
\usepackage[table]{xcolor}   % <-- 반드시 table 옵션
\usepackage{colortbl}        % \rowcolor, \cellcolor
\usepackage{booktabs,multirow,makecell,graphicx}
\usepackage{amsmath,amssymb}
\usepackage{array} 
\usepackage{adjustbox}
\usepackage{tabularx}
\usepackage[accsupp]{axessibility}  % Improves PDF readability for those with disabilities.

\usepackage{hyperref}

\usepackage{orcidlink}

\begin{document}

\title{SARIF: Segment Anything for Robust Image Forensics} 

\titlerunning{SARIF: Segment Anything for Robust Image Forensics}

\author{Dong-Hyun Moon$^{*}$\inst{1}\orcidlink{0009-0006-3215-9288} \and 
Ju-Hyeon Nam$^{*}$\inst{1}\orcidlink{0000-0002-7966-1228} \and 
Sang-Chul Lee$^{\dagger}$\inst{1}\orcidlink{0000-0002-6973-2416}}

\authorrunning{D.-H. Moon, J.-H. Nam, and S.-C. Lee}

\institute{ Department of Electrical and Computer Engineering, Inha University,\\ 100 Inha-ro, Michuhol-gu, Incheon, 22212, Republic of Korea\\ \email{\{22261207, jhnam0514\}@inha.edu, sclee@inha.ac.kr} }

\maketitle
\begingroup 
\renewcommand\thefootnote{*} 
\footnotetext{Equal contribution.} 
\renewcommand\thefootnote{$\dagger$} 
\footnotetext{Corresponding author.} 
\endgroup
\begin{abstract}
Image forgery localization remains challenging due to diverse manipulation techniques and distribution shifts. Existing recent forgery localization models achieve high accuracy on benchmarks but often struggle with cross-domain generalization and robustness. In this paper, we propose \textbf{SARIF (Segment Anything for Robust Image Forensics)}, a framework that leverages Segment Anything Model (SAM), which has a promptable architecture and generalization ability to overcome these limitations. SARIF introduces a feedback-guided mask decoder and a dual-encoder design that extracts forgery-specific information to capture forensic traces while exploiting SAM’s architecture. To localize manipulated regions, we design a block-wise prompting mechanism that derives forgery-specific cues from residual features between an adapted encoder and its frozen counterpart. These features are fused with the previous mask prompt to drive a feedback-based mask refinement process, enabling automatic forgery segmentation without manual input. Extensive experiments on standard forgery-localization benchmarks show that SARIF achieves strong average cross-dataset performance and robustness to common image corruptions. Our SARIF code is available in \href{https://github.com/Inha-CVAI/SARIF_ECCV2026}{GitHub Link}.

\end{abstract}  
\section{Introduction}
\label{sec:intro}

Large‑scale generative models \cite{kawar2023imagic, pan2023drag, yang2023paint, liu2024referring, lee2024diffusion} and widely available commercial editing software \cite{caruso2002image, barnes2009patchmatch} have made image editing ubiquitous, enabling non‑experts to create convincing image forgeries with minimal effort \cite{huang2025diffusion, meng2021sdedit, de2021distinct}. These advances have heightened concerns, as forged images fuel misinformation, trigger legal disputes, and erode public trust, undermining societal stability \cite{vaccari2020deepfakes, mubarak2023survey, ahmed2023perception}. To meet these demands, early image forgery detection relied on traditional hand‑crafted features such as illumination inconsistency traces \cite{johnson2005exposing, kee2014exposing, de2013exposing}, JPEG compression artifacts \cite{popescu2004statistical, yang2020clustering}, sensor pattern noise (SPN) \cite{lukavs2006detecting, chierchia2014bayesian, korus2016multi}, and color filter array (CFA) interpolation patterns \cite{popescu2005exposing,dirik2009image,ferrara2012image}. However, these methods focus on \textit{image-level detection} and exhibit poor generalization to unseen forgery types, often requiring specific conditions \cite{wu2019mantra, hu2020span}.

Motivated by these limitations, deep learning models, which are now the dominant paradigm across diverse vision tasks \cite{he2016deep, dosovitskiy2021an, jain2023damex, guo2022segnext, zhang2024vision}, enable \textit{pixel‑level forgery localization}. Models such as MantraNet \cite{wu2019mantra} and SPAN \cite{hu2020span} demonstrate strong performance in automatic forgery‑trace extraction. However, these approaches are often computationally intensive and have limited representational capacity, constraining generalization to unseen forgeries \cite{zhang2021multi, chen2021image}. Recently, attention mechanisms \cite{hu2018squeeze, woo2018cbam, dai2021attentional} have mitigated these issues, improving both efficiency and accuracy in UNet‑like encoder–decoder architectures \cite{zhang2021multi, hao2021transforensics, liu2022pscc}. Nevertheless, performance remains inadequate in localization accuracy, out‑of‑distribution generalization, and robustness to common corruptions.

\begin{wrapfigure}{r}{0.48\columnwidth}
  \vspace{-0.5em}
  \centering
  \includegraphics[width=\linewidth]{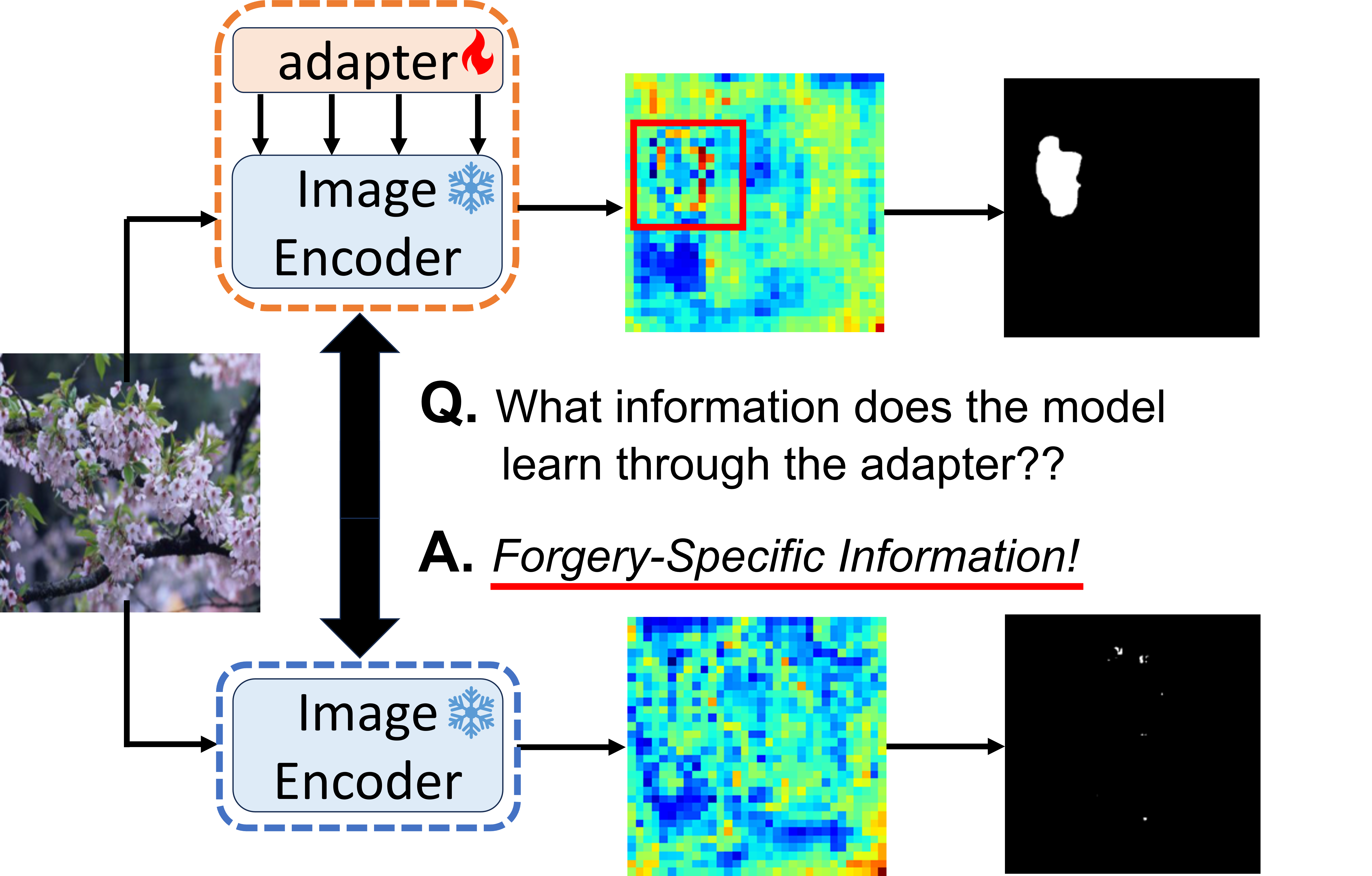}
  \caption{\textbf{\textit{Motivation}}. Without the adapter, the encoder focuses on semantic content and misses subtle manipulation artifacts, causing imprecise masks. With the adapter, it learns forgery-specific cues that guide the decoder to produce sharper and more accurate localization. Concretely, this domain gap between the adapted and the frozen encoder represents the adapter-learned forgery-specific information.}
  \vspace{-0.8em}
  \label{fig:motivation}
\end{wrapfigure}

To move beyond these limitations, we focus on the Segment Anything Model (SAM) \cite{kirillov2023segment}, a vision foundation model trained on more than a billion mask annotations across millions of images; it provides strong zero‑ and few‑shot generalization and a promptable interface. Recent image forgery localization \cite{zhang2025imdprompter, kwon2025safire, zhang2024samif} methods adapt SAM to exploit this capacity. However, many SAM-based pipelines still depend on manual prompts, which is labor-intensive, sensitive to prompt design, and often biased toward object boundaries rather than manipulation evidence, limiting practicality for forgery localization \cite{shaharabany2023autosam}. To address these issues, IMDPrompter \cite{zhang2025imdprompter} adopts multi‑view prompt learning to automate prompt generation with optimal prompt selection, yet the method remains computationally heavy and structurally complex due to multiple view‑specific encoders and auxiliary modules. Additionally, many recent studies rely solely on SAM’s final embeddings, overlooking hierarchical cues and, critically, the \textit{task‑specific information}, which SAM trained on natural images typically does not provide, needed for accurate forgery localization.

Building on these observations, we present \textit{\underline{\textbf{S}}egment \underline{\textbf{A}}nything for \underline{\textbf{R}}obust \underline{\textbf{I}}mage \underline{\textbf{F}}orensics (\underline{\textbf{SARIF}})}, a SAM-based forgery localization framework that extracts task-specific residual cues from an adapted/frozen dual-encoder pair and refines masks through automatic feedback. This design improves sensitivity to subtle manipulation traces while preserving the strong generalization ability of SAM. Additionally, we introduce a feedback‑guided mask decoder to fully exploit SAM’s lightweight decoder and promptable interface. At each feedback iteration, we fuse the previous mask prediction with forgery-specific information to yield the next refined mask, updating it via a lightweight decoder with shared weights. With feedback‑guided prompting, boundaries are progressively sharpened, spurious regions suppressed, and subtle \textit{task‑specific} cues recovered, while fully exploiting SAM’s lightweight mask decoder and preserving the base encoder’s strong generalization. Extensive experiments on standard public benchmarks show that SARIF achieves strong average performance, improved cross-domain generalization, and competitive robustness to common corruptions. The contributions of this paper can be summarized as follows:

\begin{itemize}
\item \textbf{Task-specific prompting.}
We compute block-wise task-specific features from the differences between the fine-tuned and frozen encoders at selected global-attention blocks and at the final embedding. This design lets us progressively probe what each fine-tuned layer learns during training, producing a coarse-to-fine hierarchy of task-specific information that spans broad semantic context down to fine-grained manipulation cues.

\item \textbf{Feedback-guided mask decoding with prompt--mask fusion.}
We fuse the previous mask and forgery-specific feature to form the task-specific prompt and feed it to SAM's mask decoder with shared weights to obtain the current prediction. This stage-wise refinement accumulates evidence across refinement stages, stabilizes boundaries, and fully exploits SAM's promptable design.

\item \textbf{Extensive experiments.} Across diverse benchmarks, our method shows strong localization accuracy, cross-domain generalization, and robustness under common post-processing distortions such as JPEG Compression, Gaussian Noise and Gaussian Blur.
\end{itemize}
\section{Related Work}
\label{sec:formatting}
%-------------------------------------------------------------------------
\noindent \textbf{Deep Learning Based Forgery Localization.}

\textit{\textbf{1) CNN approaches:}} Representative CNN methods include RGB Noise \cite{zhou2018learning}, which extracts SRM based noise for cross type detection, and MantraNet \cite{wu2019mantra} and SPAN \cite{hu2020span}, which use no pooling for pixel level localization. Several early models still return only bounding boxes, are computationally heavy or require extra fine tuning. To mitigate these issues, RRUNet \cite{bi2019rru} introduces feedback learning in a UNet style encoder decoder, MT-SENet \cite{zhang2021multi} adds SE blocks \cite{hu2018squeeze} with multitask learning to sharpen boundaries, and PSCCNet \cite{liu2022pscc}, HDFNet \cite{han2024hdf}, and PIMNet \cite{bai2025pim} adopt multiscale attention for adaptive fusion.

\textit{\textbf{2) Transformer approaches:}} To supply global context missing in prior CNN methods, TransForensics \cite{hao2021transforensics} uses Vision Transformer self attention \cite{dosovitskiy2021an} to capture long range dependencies and improve pixel level localization.

\textit{\textbf{3) Frequency approaches:}} Frequency features enhance cross domain generalization in various vision applications \cite{nam2023random, nam2024modality, nam2024fsda, nam2025m3fpolypsegnet++, nam2025frequency}. For forgery localization, ObjectFormer \cite{wang2022objectformer} and FBINet \cite{gu2022fbi} apply the 2D DCT to expose subtle tampering cues. M2SFormer \cite{nam2025m2sformer} fuses multiscale and multispectral signals for better generalization and corruption robustness, while DNet \cite{yang2024d}, AFENet \cite{xu2024image}, and EITLNet \cite{guo2024effective} further strengthen generalization using a Haar wavelet transform, a 2D discrete Fourier transform, and a high pass noise filter with RGB inputs, respectively.

\textit{\textbf{4) Segment Anything approaches:}} Vision foundation models such as SAM \cite{kirillov2023segment, ravi2024sam} provide promptable mask generation with strong zero-shot and few-shot generalization, motivating their use in forgery localization. IMDPrompter \cite{zhang2025imdprompter} freezes the encoder and learns multi-view prompts with a noise-oriented CNN and optimal selection, while SAMIF \cite{zhang2024samif} augments the image encoder with adapters and an SRM branch to emphasize high-frequency cues. SAFIRE \cite{kwon2025safire} reframes forgery localization as source-region partitioning: it performs point-prompted source-region segmentation and aggregates predictions from a grid of automatically generated points to recover source partitions and forged regions. Nevertheless, many SAM-based pipelines remain prompt-dependent or reduce SAM to a single final embedding, leaving hierarchical signals and task-specific manipulation cues underexplored. These limitations motivate our design, which injects hierarchical cues through dual-encoder and automatic feedback refinement while preserving SAM’s promptable decoder.

\noindent \textbf{Segment Anything for Computer Vision.} The Segment Anything Model (SAM) \cite{kirillov2023segment, ke2023segment, ravi2024sam} comprises an image encoder and a promptable, lightweight mask decoder that accepts points, boxes, or masks as prompts and generalizes in zero-shot across diverse distributions. Enabled by this generalization, SAM is increasingly adopted as a downstream backbone across tasks. In medical imaging, MedSAM \cite{ma2024segment}, AutoSAM \cite{shaharabany2023autosam}, SAM‑Med2D \cite{wu2023medical}, and SAM‑Med3D \cite{zhu2024medical} extend SAM to diverse 2D and 3D modalities, delivering versatile organ and lesion segmentation with minimal prompting. In camouflaged object detection, COD‑SAM \cite{gao2025cod} and SAM‑COD+ \cite{chen2024sam} employ adapters, distillation, and task aware prompting to sharpen subtle boundary cues. For forgery localization, \textit{\textbf{SARIF}} combines parameter-efficient adaptation, dual-encoder residual extraction, and feedback-guided decoding to inject task-specific cues without relying on manual prompts or a single final embedding.

\begin{figure*}[t]
  \centering
  \includegraphics[width=\textwidth]{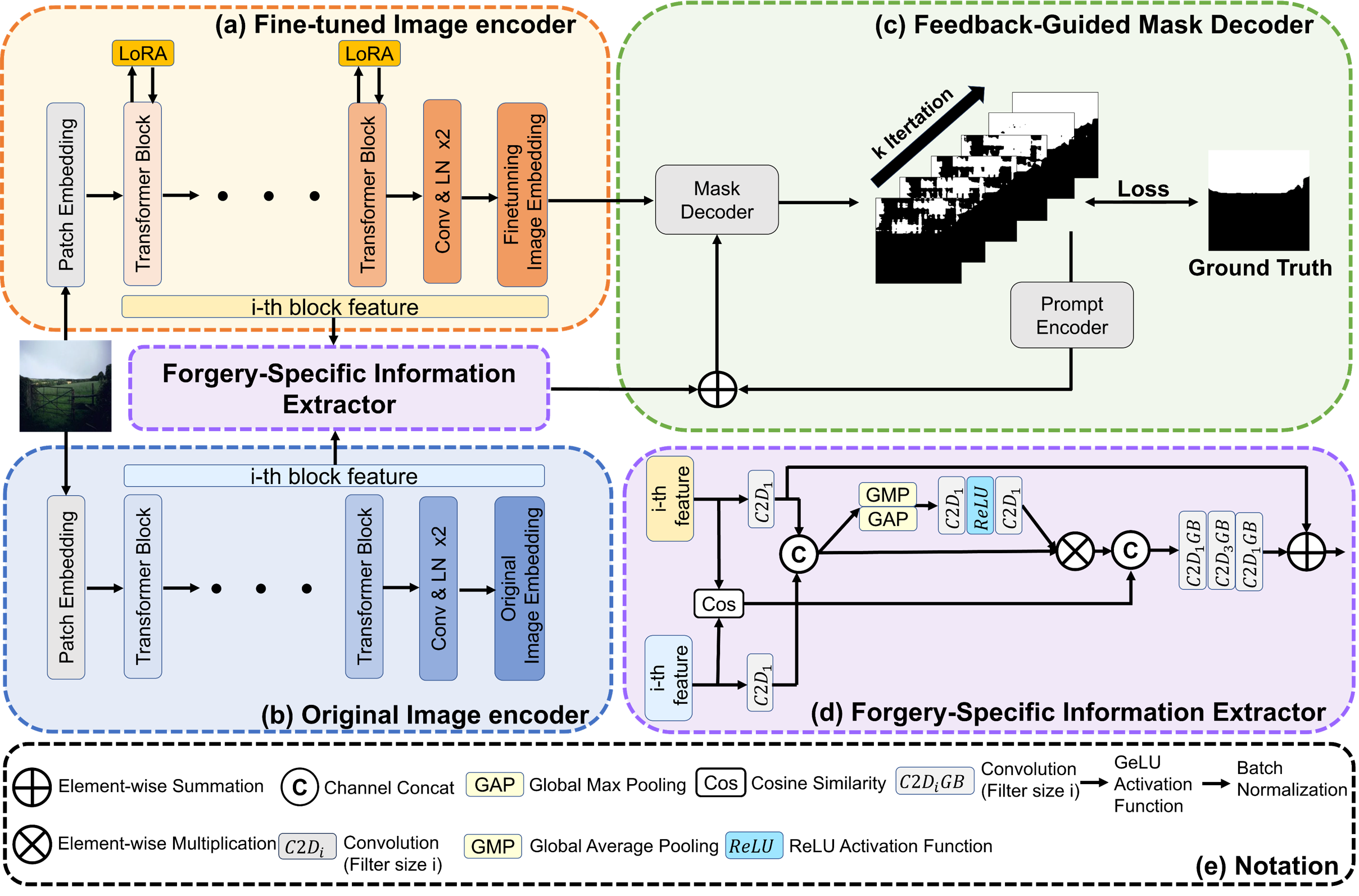}
  \caption{The overall architecture of the proposed \textbf{SARIF}. (a) Fine-tuned SAM Image Encoder. (b) Original SAM Image Encoder. (c) Feedback-Guided Mask Decoder. (d) Forgery-Specific Information Extractor. (e) Notation description used in this paper.}
  \label{fig:overall_sarif}
\end{figure*}
\section{Proposed Method}
\subsection{Overall Architecture}
To fully leverage SAM’s hierarchical representations and promptable mask decoder for forgery localization, we propose our overall framework, \textbf{S}egment \textbf{A}nything for \textbf{R}obust \textbf{I}mage \textbf{F}orensics (SARIF), illustrated in Fig. \ref{fig:overall_sarif}. SARIF comprises three modules: (\textit{1) a dual SAM image encoder}, \textit{(2) the Forgery-Specific Information Extractor (FSIE; Sec. \ref{ssec_forgery_specific_information_extraction})}, and \textit{(3) the Feedback-Guided Mask Decoder (FGMD; Sec. \ref{ssec_iterative_mask_refine_decoder})}. As illustrated in Fig. \ref{fig:overall_sarif} (a) and (b), we insert LoRA adapters \cite{hu2022lora} into one image-encoder branch and update only these adapters and mask decoder for forgery localization, while keeping the backbone weights frozen. At each selected transformer block, FSIE computes block-wise residual cues between the adapted and frozen branches to distill forgery-specific information, which is then fused with the previous mask prompt to produce the next task-specific prompt. At each refinement step, the decoder takes the fine-tuned SAM embedding together with this prompt to predict and refine the mask. The final mask is obtained by leveraging a task-specific prompt derived from the fine-tuned and original embeddings. By capturing what SAM learns through LoRA and fully using SAM’s promptable design and lightweight mask decoder, the framework precisely adapts SAM to the forgery localization task.

\subsection{Forgery Specific Information Extractor}
\label{ssec_forgery_specific_information_extraction}
\noindent \textit{Motivation.} We observe that a single branch adapter setting is parameter efficient but provides no explicit understanding of what the adapter learns for the forgery specific task, limiting sensitivity to subtle manipulations. Motivated by primate biological vision theories \cite{kuffler1953discharge, hurvich1957opponent, mcmahon2004classical}, where opponent channels and center surround circuitry encode relative contrast rather than absolute intensity, we pair the pretrained SAM encoder with its LoRA adapted counterpart and compute blockwise residuals at transformer block to form compact forgery specific cues. To realize this, we introduce the \textit{\textbf{Forgery Specific Information Extractor (FSIE)}}, which aggregates these features with a lightweight residual block to produce low cost features that preserve manipulation sensitivity while keeping the dual branch design practical. This design is also inspired by HQ-SAM \cite{ke2023segment}, which taps features immediately after global-attention blocks; accordingly, we therefore compute residual cues only at global-attention blocks, where long-range interactions are explicitly enabled, capturing global inconsistencies from forgeries with minimal overhead. FSIE operates in three stages: \textit{1) Feature Similarity Calculation}, \textit{2) Feature Enhancement}, and \textit{3) Fusion}.

\noindent \textbf{Feature Similarity Calculation.} Let $\mathbf{O}^{t}, \mathbf{F}^{t} \in \mathbb{R}^{C \times H \times W}$ denote the features extracted at iteration $t$ from the original and fine-tuned SAM encoders, respectively. We first compute the cosine similarity map \(\mathbf{cos}^t\) between the input feature maps, which is utilized later in subsequent stages. 
\begin{align}
\mathbf{cos}^t_{ij} \;=\; 
\frac{\langle {\mathbf{O}}^t_{:,ij}, {\mathbf{F}}^t_{:,ij} \rangle}
{\lVert {\mathbf{O}}^t_{:,ij} \rVert_2 \, \lVert {\mathbf{F}}^t_{:,ij} \rVert_2},
\quad \mathbf{cos}^t \in \mathbb{R}^{1 \times H \times W}.
\end{align}

Where \(\langle \cdot,\cdot \rangle\) inner product, \(\lVert \cdot \rVert_2\) \(\ell_2\)-norm. Then we project both branches to $C^{'} < C$ using $1 \times 1$ convolution, which reduces computation and aligns the two branches for subsequent enhancement and fusion.

\noindent \textbf{Feature Enhancement.}
To estimate the channel importance within the feature map of each stage $t$, we apply a channel attention mechanism. Specifically, we apply global average pooling and global max pooling over the spatial dimensions to obtain two channel-wise descriptors, processed by a two 1×1 convolution layer with an intervening ReLU.
\begin{align}
\mathbf{C}^t &= \operatorname{Cat}\!\left(\mathbf{F}^t, \mathbf{O}^t\right) \in \mathbb{R}^{2C' \times H \times W}, \\
\mathbf{g}_{\max}^t &= \operatorname{Conv}_{1\times1}\!\bigl(
\phi(
\operatorname{Conv}_{1\times1}(
\operatorname{GMP}(\mathbf{C}^t)
)
)
\bigr), \\
\mathbf{g}_{\mathrm{avg}}^t &= \operatorname{Conv}_{1\times1}\!\bigl(
\phi(
\operatorname{Conv}_{1\times1}(
\operatorname{GAP}(\mathbf{C}^t)
)
)
\bigr).
\end{align}

Denote \(\phi\) is ReLU activation function. Then aggregate their outputs, and apply a sigmoid function to generate channel-wise attention weights. These weights are multiplied with the original feature map to emphasize informative channels while suppressing less relevant ones.
\begin{align}
\mathbf{G}^t &= \sigma\!\left( \mathbf{g}_{\max}^t + \mathbf{g}_{\mathrm{avg}}^t \right), \\
\mathbf{T}_{\mathrm{ca}}^t &= \mathbf{C}^t \odot \mathbf{G}^t \in \mathbb{R}^{2C' \times H \times W}.
\end{align}

Denote $\sigma(\cdot)$ is the sigmoid function, $\odot$ denotes channel-wise reweighting with spatial broadcasting. 

\noindent \textbf{Fusion.}
We use the previously computed cosine similarity map as a hint map that guides the network toward forgery-specific information and fuse it with the channel-attended feature map. 
\begin{align}
\mathbf{Z}^t &= \operatorname{Cat}\!\left(\mathbf{T}_{\mathrm{ca}}^t, \mathbf{cos}^t\right) \in \mathbb{R}^{(2C' + 1) \times H \times W}.
\end{align}
To extract forgery-specific information, we employ a lightweight convolutional layer, GeLU activation, batch normalization with a skip connection. For computational efficiency, we reduce the channel dimension to $r < C'$, apply a $3 \times 3$ convolution in this reduced space, and then project the features back to the original embedding dimension.
\begin{align}
\mathbf{T}_{\mathrm{fs}}^t
= \mathbf{F}^t
\oplus \mathbf{CGB}_{1\times1}^{C'}\!\Big(
    \mathbf{CGB}_{3\times3}\!\big(
        \mathbf{CGB}_{1\times1}^{r}(\mathbf{Z}^t)
    \big)
\Big),
\end{align}

We extract forgery-specific features at the global-attention blocks with indices \{5, 11, 17, 23\}, and at final embedding. These stage-wise cues are sequentially injected into the Feedback-Guided Mask Decoder together with the image embedding and previous mask prompt to progressively refine masks.

\begin{figure*}[t]
  \centering
  \includegraphics[width=\textwidth]{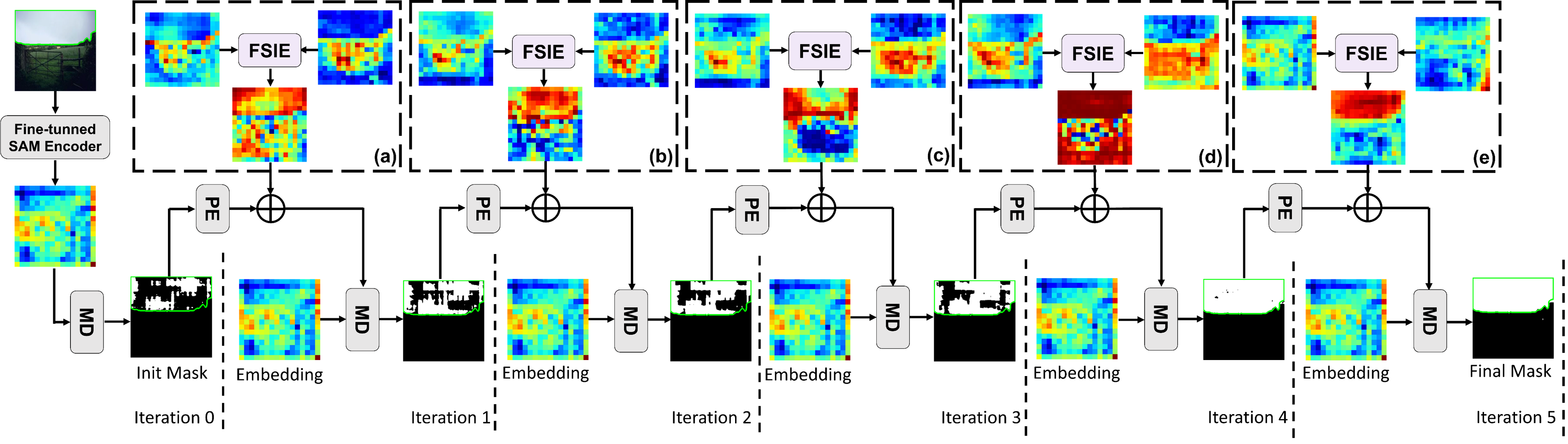}
  \caption{\textbf{More details about Feedback-Guided Mask Decoder.} (a$-$e) show features from the 5th, 11th, 17th, and 23rd blocks and the final embedding. We also visualize forgery-specific information maps produced by FSIE. For each FSIE input pair, the left feature map comes from the fine-tuned encoder and the right one comes from the original frozen encoder. At each refinement stage, the previous mask is passed to the prompt encoder (PE) to generate a mask prompt, which is fused with the forgery-specific information produced by FSIE to form a task-specific prompt. This task-specific prompt, together with the fine-tuned image embedding, is fed into the mask decoder (MD) to progressively refine the mask. \textcolor{green}{\textbf{Green}} lines denote the boundaries of the ground truth.}
  \label{fig:Feedback}
  \vspace{-1.2em}
\end{figure*}

\subsection{Feedback-Guided Mask Decoder}
\label{ssec_iterative_mask_refine_decoder}
\noindent \textit{Motivation.} SAM provides a promptable architecture that can be reused at inference, which we leverage to progressively recover fine mask detail \cite{kirillov2023segment}. However, most automatic SAM based forgery localization methods do not reuse previous predictions as feedback \cite{zhang2024samif, zhang2025imdprompter}. We therefore design a \textit{\textbf{Feedback-Guided Mask Decoder (FGMD)}} that, at each refinement stage, fuses the previous mask after SAM prompt encoding with FSIE cues to form a new prompt and refine boundaries in a coarse to fine hierarchy, consistent with human vision system where center and surround interactions encode relative contrast and recurrent processing sharpens perception \cite{hupe1998cortical, lamme2000distinct, kirchberger2021essential}. This design also accords with evidence that iterative refinement improves segmentation accuracy and boundary fidelity at modest computational cost. FGMD operates in two stages: \textit{1) Forgery-specific Prompt Generation}, and \textit{2) Iterative Mask Refinement}.

\noindent \textbf{Forgery-specific Prompt Generation.}
Prior to mask refinement, an initial mask is required to serve as the reference for subsequent updates. We define an initial mask as the mask decoder’s output given only the fine-tuned SAM image embeddings, without external prompts. Finally, the previous mask is defined as:
\begin{align}
{\mathbf{M}}^{(t-1)} \;=\;
\begin{cases}
\mathbf{None} & \text{if } t=1,\\
\mathbf{M}^{(t-1)} & \text{if } t>1,
\end{cases}
\end{align}
For refinement step $t$, we encode the previous mask $\mathbf{M}^{t - 1}$ with the prompt encoder and fuse the resulting mask prompt with the $t$-th forgery-specific cue extracted by FSIE.
\begin{align}
\mathbf{Mask}_{prompt}^{t}\;=\;
\mathrm{PromptEncoder}\!\big({{\mathbf{M}}^{t-1}}\big).
\end{align}
\vspace{-1.2em}
\begin{align}
\mathbf{T}_{\mathbf{fs}}^{t} \;=\; {\mathrm{FSIE}}\!\big(\mathbf{F}^{t},\, {\mathbf{O}}^{t}\big).
\end{align}
%In particular, the mask-prompt branch of the PromptEncoder is implemented as a lightweight three-layer convolutional network, which embeds an input mask prompt into SAM’s prompt feature space.
 Where PromptEncoder denotes SAM’s prompt encoder. To generate the task-specific prompt, the task-specific features are merged with the mask prompt via element-wise summation.
\begin{align}
\mathbf{T}_{\mathbf{pt}}^{t}\;=\;(\mathbf{Mask}_{Prompt}^{t}\oplus\mathbf{T}_{\mathbf{fs}}^{t}\big).
\end{align}
\noindent \textbf{Stage-wise Mask Refinement.}
Let $\mathbf{T}_{\mathrm{pt}}^t$ denote the task-specific prompt at refinement stage $t$.
The SAM mask decoder $\mathcal{D}$ takes the final fine-tuned image embedding
$\mathbf{E}_{\mathrm{fine}}$ and $\mathbf{T}_{\mathrm{pt}}^t$ as inputs and produces the refined prediction:
\begin{equation}
    \mathbf{M}^{t} = \sigma\!\left(
    \mathcal{D}\!\left(\mathbf{E}_{\mathrm{fine}}, \mathbf{T}_{\mathrm{pt}}^{t}\right)
    \right).
\end{equation}
Here, $\sigma(\cdot)$ denotes the sigmoid function.
The predicted mask is thresholded and passed to the next refinement stage as a mask prompt.
Each refinement stage is supervised with the BCE loss against the ground-truth mask:
\begin{equation}
    \mathcal{L}_{\mathrm{BCE}}
    =
    \sum_{t=0}^{5}
    \mathrm{BCE}\!\left(\mathbf{M}^{t}, \mathbf{Y}\right),
\end{equation}
where $\mathbf{Y}$ denotes the ground-truth mask. This design injects task cues through the prompt path, preserves accumulated spatial context through the mask path, and performs progressive refinement with a lightweight decoder. Further details of the refinement are shown in Figure \ref{fig:Feedback}.

\section{Experiment Results}
\subsection{Experiment Settings}
To evaluate model generalization, we train under CASIAv2 \cite{pham2019hybrid}, then assess cross domain performance on seven external datasets: CASIAv1 \cite{Dong2013}, Columbia \cite{hsu06crfcheck}, IMD2020 \cite{Novozamsky_2020_WACV}, CoMoFoD \cite{tralic2013comofod}, In the Wild \cite{huh2018fighting}, MISD \cite{kadam2021multiple} and DIS25k \cite{tahir2024deep}. We emphasize that none of these benchmarks are used for training, in line with the evaluation protocol followed by the recent state of the art M2SFormer \cite{nam2025m2sformer}. Because of page limits, we provide detailed dataset summaries in Table \ref{tab:dataset_summary} and the exact split definitions for each training scheme in Appendix Section \ref{appendix_dataset_descriptions}. To measure performance, we report Dice Similarity Coefficient (DSC) and mean Intersection over Union (mIoU), which are established metrics for segmentation; formal definitions and implementation details appear in Appendix Section \ref{appendix_metric_descriptions}.

We compare the proposed \textbf{SARIF (Ours)} against thirteen representative forgery localization models, including eight convolutional-based approaches (UNet \cite{ronneberger2015u}, ManTraNet \cite{wu2019mantra}, RRU-Net \cite{bi2019rru}, TransForensics \cite{hao2021transforensics}, FBI-Net \cite{gu2022fbi}, MT-SENet \cite{zhang2021multi}, MVSS-Net \cite{dong2022mvss}, PIM-Net \cite{bai2025pim}), two Transformer-based approaches (EITLNet \cite{guo2024effective}, M2SFormer \cite{nam2025m2sformer}), and four SAM-based approaches (SAM \cite{kirillov2023segment} used without prompts, AutoSAM \cite{shaharabany2023autosam}, IMDPrompter \cite{zhang2025imdprompter}, SAFIRE \cite{kwon2025safire}). For a fair comparison, all methods are trained on the same training set with matched preprocessing and hyperparameters where applicable, and evaluated on an identical held out test set to assess generalization. We report the mean over five-fold cross-validation for reliability. In all tables, \textcolor{red}{\textbf{{Red}}} and \textcolor{blue}{\textbf{\textit{Blue}}} denote the best and second best results.

\begin{table*}[t]
    \centering
    \caption{Segmentation results (DSC) on CASIAv2 \cite{pham2019hybrid} training scheme. $(\cdot)$ denotes the standard deviations of five-fold cross-validation experiment results.}
        \tiny
        \renewcommand{\arraystretch}{1.25} % Tighter
        \resizebox{\textwidth}{!}{%
        \begin{tabular}{c|c|c|c|c|c|c|c|c|c}
            \hline
            \cellcolor[gray]{.9} &
            \cellcolor[gray]{.9} &
            \multicolumn{1}{c|}{\cellcolor[gray]{.9}Seen Domain} &
            \multicolumn{7}{c}{\cellcolor[gray]{.9}Unseen Domain} \\
            \cline{3-10}\cline{3-10}

            \cellcolor[gray]{.9} &
            \cellcolor[gray]{.9} &
            \multicolumn{1}{c|}{\cellcolor[gray]{.9}CASIAv2 \cite{pham2019hybrid}} &
            \multicolumn{1}{c|}{\cellcolor[gray]{.9}DIS25k \cite{tahir2024deep}} &
            \multicolumn{1}{c|}{\cellcolor[gray]{.9}CASIAv1 \cite{Dong2013}} &
            \multicolumn{1}{c|}{\cellcolor[gray]{.9}Columbia \cite{hsu06crfcheck}} &
            \multicolumn{1}{c|}{\cellcolor[gray]{.9}IMD2020 \cite{Novozamsky_2020_WACV}} &
            \multicolumn{1}{c|}{\cellcolor[gray]{.9}CoMoFoD \cite{tralic2013comofod}} &
            \multicolumn{1}{c|}{\cellcolor[gray]{.9}In the Wild \cite{huh2018fighting}} &
            \multicolumn{1}{c}{\cellcolor[gray]{.9}MSID \cite{kadam2021multiple}} \\
            \cline{3-10}\cline{3-10}

            \multirow{-3}{*}{\cellcolor[gray]{.9}Type} &
            \multirow{-3}{*}{\cellcolor[gray]{.9}Method} &
            \cellcolor[gray]{0.9}DSC &
            \cellcolor[gray]{0.9}DSC &
            \cellcolor[gray]{0.9}DSC &
            \cellcolor[gray]{0.9}DSC &
            \cellcolor[gray]{0.9}DSC &
            \cellcolor[gray]{0.9}DSC &
            \cellcolor[gray]{0.9}DSC &
            \cellcolor[gray]{0.9}DSC \\
            \hline

            \rowcolor{red!10}
            & UNet \cite{ronneberger2015u}
            & 32.3 (10.9)
            & 8.7 (1.2)
            & 25.0 (1.9)
            & 19.8 (2.5)
            & 14.9 (1.0)
            & 12.2 (1.0)
            & 18.6 (1.4)
            & 47.3 (2.5) \\

            \rowcolor{red!10}
            & MantraNet \cite{wu2019mantra}
            & 18.8 (8.0)
            & 12.7 (0.8)
            & 19.8 (1.0)
            & 25.0 (1.8)
            & 14.1 (0.5)
            & 11.2 (0.6)
            & 18.2 (1.1)
            & 30.7 (3.5) \\

            \rowcolor{red!10}
            & RRUNet \cite{bi2019rru}
            & 21.8 (10.4)
            & 10.8 (2.2)
            & 26.7 (2.4)
            & 21.3 (7.2)
            & 14.5 (1.9)
            & 13.7 (2.2)
            & 18.3 (4.2)
            & 32.8 (3.4) \\

            \rowcolor{red!10}
            & TransForensic \cite{hao2021transforensics}
            & 40.1 (15.7)
            & 32.3 (2.9)
            & 44.2 (1.6)
            & 35.9 (5.7)
            & 27.2 (2.1)
            & 21.7 (2.0)
            & 31.9 (3.8)
            & 60.0 (1.8) \\

            \rowcolor{red!10}
            & FBINet \cite{gu2022fbi}
            & 35.3 (14.3)
            & 26.3 (2.3)
            & 37.5 (2.1)
            & 18.2 (2.7)
            & 24.2 (1.6)
            & 22.3 (0.8)
            & 25.0 (2.7)
            & 48.0 (2.2) \\

            \rowcolor{red!10}
            & MT-SENet \cite{zhang2021multi}
            & 19.9 (9.9)
            & 7.8 (1.0)
            & 18.4 (1.6)
            & 9.4 (1.1)
            & 11.4 (1.3)
            & 10.6 (2.1)
            & 10.6 (2.1)
            & 22.3 (2.5) \\

            \rowcolor{red!10}
            \multirow{-7}{*}{\rotatebox[origin=c]{90}{\textbf{Convolution}}}
            & MVSSNet \cite{dong2022mvss}
            & 31.2 (13.4)
            & 24.9 (3.6)
            & 36.6 (1.6)
            & 33.8 (3.7)
            & 22.8 (2.4)
            & 17.2 (1.2)
            & 27.0 (3.1)
            & 53.9 (3.6) \\
            \hline

            \rowcolor{orange!10}
            & PIMNet \cite{bai2025pim}
            & 55.8 (15.1)
            & 37.5 (2.4)
            & 49.7 (0.8)
            & 32.5 (5.2)
            & 29.6 (2.7)
            & 24.7 (1.6)
            & 31.2 (2.5)
            & 61.1 (0.9) \\

            \rowcolor{orange!10}
            & EITLNet \cite{guo2024effective}
            & 54.0 (14.7)
            & 30.8 (2.8)
            & 52.9 (1.7)
            & 28.0 (4.6)
            & 25.3 (2.6)
            & 18.1 (1.8)
            & 24.3 (3.6)
            & 58.8 (1.8) \\

            \rowcolor{orange!10}
            \multirow{-3}{*}{\rotatebox[origin=c]{90}{\textbf{Trans.}}}
            & M2SFormer \cite{nam2025m2sformer}
            & \textcolor{blue}{\textbf{\textit{58.8 (12.8)}}}
            & 38.5 (2.4)
            & \textcolor{blue}{\textbf{\textit{58.4 (0.7)}}}
            & 42.4 (5.8)
            & 32.6 (2.2)
            & 24.9 (1.3)
            & 35.0 (1.8)
            & \textbf{{\textcolor{red}{69.1 (0.7)}}} \\
            \hline

            \rowcolor{green!10}
            & SAM \cite{kirillov2023segment}
            & 27.1 (12.2)
            & 18.7 (3.4)
            & 33.4 (6.4)
            & 24.6 (11.3)
            & 22.3 (2.7)
            & 19.7 (3.0)
            & 29.6 (8.1)
            & 31.3 (8.1) \\

            \rowcolor{green!10}
            & AutoSAM \cite{shaharabany2023autosam}
            & 49.0 (19.5)
            & 31.4 (1.4)
            & 45.1 (4.1)
            & 28.2 (9.0)
            & 34.0 (3.0)
            & 23.8 (1.7)
            & 36.5 (12.0)
            & 61.3 (5.5) \\

            \rowcolor{green!10}
            & IMDPromter \cite{zhang2025imdprompter}
            & 32.5 (16.9)
            & 28.6 (2.2)
            & 46.1 (2.8)
            & 33.9 (6.6)
            & 28.0 (2.2)
            & 23.0 (1.1)
            & 33.7 (3.1)
            & 47.4 (2.7) \\

            \rowcolor{green!10}
            & SAFIRE \cite{kwon2025safire}
            & 56.8 (10.4)
            & \textbf{{\textcolor{red}{50.7 (0.9)}}}
            & \textbf{{\textcolor{red}{61.6 (0.9)}}}
            & \textcolor{blue}{\textbf{\textit{53.1 (4.9)}}}
            & \textbf{{\textcolor{red}{53.1 (0.6)}}}
            & \textcolor{blue}{\textbf{\textit{39.5 (1.0)}}}
            & \textbf{{\textcolor{red}{63.3 (2.2)}}}
            & 48.3 (1.5) \\

            \rowcolor{green!10}
            \multirow{-5}{*}{\rotatebox[origin=c]{90}{\textbf{SAM}}}
            & \textbf{SARIF (Ours)}
            & \textbf{{\textcolor{red}{63.1 (12.9)}}}
            & \textcolor{blue}{\textbf{\textit{47.8 (2.5)}}}
            & \textcolor{blue}{\textbf{\textit{58.4 (1.3)}}}
            & \textbf{{\textcolor{red}{58.4 (3.4)}}}
            & \textcolor{blue}{\textbf{\textit{48.4 (1.7)}}}
            & \textbf{{\textcolor{red}{65.4 (2.0)}}}
            & \textcolor{blue}{\textbf{\textit{55.7 (3.5)}}}
            & \textcolor{blue}{\textbf{\textit{66.2 (1.9)}}} \\
            \hline
        \end{tabular}}
    \vspace{-0.5em}
    \label{tab:placeholder_dsc}
\end{table*}

% =========================
% Table 2: mIoU only
% =========================
\begin{table}[t]
    \centering
    \caption{Segmentation results (mIoU) on CASIAv2 \cite{pham2019hybrid} training scheme. $(\cdot)$ denotes the standard deviations of five-fold cross-validation experiment results.}
        \tiny
        \renewcommand{\arraystretch}{1.25} % Tighter
        \resizebox{\textwidth}{!}{%
        \begin{tabular}{c|c|c|c|c|c|c|c|c|c}
            \hline
            \cellcolor[gray]{.9} &
            \cellcolor[gray]{.9} &
            \multicolumn{1}{c|}{\cellcolor[gray]{.9}Seen Domain} &
            \multicolumn{7}{c}{\cellcolor[gray]{.9}Unseen Domain} \\
            \cline{3-10}\cline{3-10}

            \cellcolor[gray]{.9} &
            \cellcolor[gray]{.9} &
            \multicolumn{1}{c|}{\cellcolor[gray]{.9}CASIAv2 \cite{pham2019hybrid}} &
            \multicolumn{1}{c|}{\cellcolor[gray]{.9}DIS25K \cite{tahir2024deep}} &
            \multicolumn{1}{c|}{\cellcolor[gray]{.9}CASIAv1 \cite{Dong2013}} &
            \multicolumn{1}{c|}{\cellcolor[gray]{.9}Columbia \cite{hsu06crfcheck}} &
            \multicolumn{1}{c|}{\cellcolor[gray]{.9}IMD2020 \cite{Novozamsky_2020_WACV}} &
            \multicolumn{1}{c|}{\cellcolor[gray]{.9}CoMoFoD \cite{tralic2013comofod}} &
            \multicolumn{1}{c|}{\cellcolor[gray]{.9}In the Wild \cite{huh2018fighting}} &
            \multicolumn{1}{c}{\cellcolor[gray]{.9}MSID \cite{kadam2021multiple}} \\
            \cline{3-10}\cline{3-10}

            \multirow{-3}{*}{\cellcolor[gray]{.9}Type} &
            \multirow{-3}{*}{\cellcolor[gray]{.9}Method} &
            \cellcolor[gray]{0.9}mIoU &
            \cellcolor[gray]{0.9}mIoU &
            \cellcolor[gray]{0.9}mIoU &
            \cellcolor[gray]{0.9}mIoU &
            \cellcolor[gray]{0.9}mIoU &
            \cellcolor[gray]{0.9}mIoU &
            \cellcolor[gray]{0.9}mIoU &
            \cellcolor[gray]{0.9}mIoU \\
            \hline

            \rowcolor{red!10}
            & UNet \cite{ronneberger2015u}
            & 25.8 (9.2)
            & 5.5 (0.8)
            & 19.5 (1.8)
            & 12.2 (1.7)
            & 9.7 (0.6)
            & 7.6 (0.7)
            & 11.9 (1.0)
            & 34.8 (2.2) \\

            \rowcolor{red!10}
            & MantraNet \cite{wu2019mantra}
            & 11.9 (5.7)
            & 7.4 (0.6)
            & 12.1 (0.8)
            & 14.9 (1.2)
            & 8.2 (0.3)
            & 6.5 (0.4)
            & 10.6 (0.8)
            & 19.2 (2.5) \\

            \rowcolor{red!10}
            & RRUNet \cite{bi2019rru}
            & 15.8 (8.4)
            & 6.9 (1.5)
            & 18.9 (1.7)
            & 13.9 (5.3)
            & 9.4 (3.3)
            & 9.0 (1.7)
            & 11.7 (3.0)
            & 21.4 (2.5) \\

            \rowcolor{red!10}
            & TransForensic \cite{hao2021transforensics}
            & 32.0 (13.9)
            & 24.1 (2.5)
            & 35.0 (1.7)
            & 25.0 (4.6)
            & 19.1 (1.6)
            & 14.3 (1.5)
            & 22.4 (3.0)
            & 46.5 (2.1) \\

            \rowcolor{red!10}
            & FBINet \cite{gu2022fbi}
            & 29.2 (12.8)
            & 20.3 (1.9)
            & 30.8 (1.9)
            & 11.7 (2.1)
            & 17.5 (1.3)
            & 15.2 (0.5)
            & 17.8 (2.1)
            & 35.1 (2.0) \\

            \rowcolor{red!10}
            & MT-SENet \cite{zhang2021multi}
            & 14.6 (7.6)
            & 4.8 (0.7)
            & 12.8 (1.3)
            & 5.3 (0.6)
            & 7.2 (1.0)
            & 6.5 (1.5)
            & 6.5 (1.5)
            & 14.0 (1.8) \\

            \rowcolor{red!10}
            \multirow{-7}{*}{\rotatebox[origin=c]{90}{\textbf{Convolution}}}
            & MVSSNet \cite{dong2022mvss}
            & 23.4 (11.1)
            & 17.4 (2.9)
            & 27.3 (1.5)
            & 23.3 (3.0)
            & 15.2 (1.8)
            & 10.9 (0.9)
            & 18.3 (2.2)
            & 17.4 (2.9) \\
            \hline

            \rowcolor{orange!10}
            & PIMNet \cite{bai2025pim}
            & 48.5 (14.6)
            & 30.1 (2.3)
            & 42.2 (1.0)
            & 23.1 (4.3)
            & 22.2 (2.3)
            & 16.8 (1.4)
            & 22.9 (2.2)
            & 48.2 (0.9) \\

            \rowcolor{orange!10}
            & EITLNet \cite{guo2024effective}
            & 47.9 (14.7)
            & 25.6 (2.6)
            & 46.5 (1.5)
            & 20.9 (4.0)
            & 19.7 (2.2)
            & 12.4 (1.4)
            & 19.0 (3.1)
            & 45.9 (1.8) \\

            \rowcolor{orange!10}
            \multirow{-3}{*}{\rotatebox[origin=c]{90}{\textbf{Trans.}}}
            & M2SFormer \cite{nam2025m2sformer}
            & \textcolor{blue}{\textbf{\textit{50.8 (12.8)}}}
            & 31.3 (2.3)
            & \textcolor{blue}{\textbf{\textit{50.1 (0.6)}}}
            & 32.4 (5.3)
            & 24.9 (1.9)
            & 16.8 (1.0)
            & 27.4 (1.6)
            & \textbf{{\textcolor{red}{56.9 (0.8)}}} \\
            \hline

            \rowcolor{green!10}
            & SAM \cite{kirillov2023segment}
            & 21.0 (10.5)
            & 14.0 (3.1)
            & 26.9 (6.1)
            & 15.9 (8.0)
            & 16.1 (2.1)
            & 12.8 (2.2)
            & 20.3 (6.9)
            & 20.0 (5.7) \\

            \rowcolor{green!10}
            & AutoSAM \cite{shaharabany2023autosam}
            & 41.7 (19.0)
            & 24.6 (1.6)
            & 38.9 (4.4)
            & 18.9 (7.6)
            & 25.6 (2.3)
            & 16.3 (1.4)
            & 25.6 (11.0)
            & 46.1 (5.9) \\

            \rowcolor{green!10}
            & IMDPromter \cite{zhang2025imdprompter}
            & 25.9 (14.9)
            & 21.3 (2.0)
            & 37.9 (3.3)
            & 22.9 (4.5)
            & 19.7 (2.0)
            & 15.4 (0.8)
            & 23.4 (2.9)
            & 33.1 (2.5) \\

            \rowcolor{green!10}
            & SAFIRE \cite{kwon2025safire}
            & 45.5 (10.9)
            & \textcolor{blue}{\textbf{\textit{40.9 (1.2)}}}
            & 49.6 (0.9)
            & \textcolor{blue}{\textbf{\textit{40.5 (4.7)}}}
            & \textbf{{\textcolor{red}{41.1 (0.6)}}}
            & \textcolor{blue}{\textbf{\textit{26.4 (0.8)}}}
            & \textbf{{\textcolor{red}{51.0 (2.2)}}}
            & 33.9 (1.3) \\

            \rowcolor{green!10}
            \multirow{-5}{*}{\rotatebox[origin=c]{90}{\textbf{SAM}}}
            & \textbf{SARIF (Ours)}
            & \textbf{{\textcolor{red}{56.7 (13.3)}}}
            & \textbf{{\textcolor{red}{41.5 (2.4)}}}
            & \textbf{{\textcolor{red}{52.0 (1.1)}}}
            & \textbf{{\textcolor{red}{49.7 (3.1)}}}
            & \textcolor{blue}{\textbf{\textit{40.4 (1.6)}}}
            & \textbf{{\textcolor{red}{55.8 (2.0)}}}
            & \textcolor{blue}{\textbf{\textit{48.6 (3.6)}}}
            & \textcolor{blue}{\textbf{\textit{53.9 (2.0)}}} \\
            \hline
        \end{tabular}}
    \vspace{-0.5em}
    \label{tab:placeholder_miou}
\end{table}
\subsection{Implementation Details}
\noindent\textbf{Training Settings.} All models were optimized with Adam \cite{kingma2014adam} starting from an initial learning rate of $10^{-4}$, and the learning rate was annealed to $10^{-6}$ using a cosine schedule \cite{loshchilov2016sgdr}. We trained for 100 epochs with a batch size of 32. Because the datasets provide images at heterogeneous native resolutions, we resized all inputs to $256 \times 256$ for the unified setting. Additionally, to ensure a controlled comparison, we used identical preprocessing, optimizer, and scheduler configurations for all models across all experiments.

\noindent\textbf{Hyperparameters of SARIF.}
We adopt SAM’s ViT-L pretrained weights and set LoRA rank as 32 in this paper. We extract intermediate embeddings at global attention block \{5, 11, 17, 23\} and the final embedding from both the fine-tuned SAM image encoder and the original frozen encoder.

\begin{figure*}[t]
  \centering
  \includegraphics[width=\linewidth]{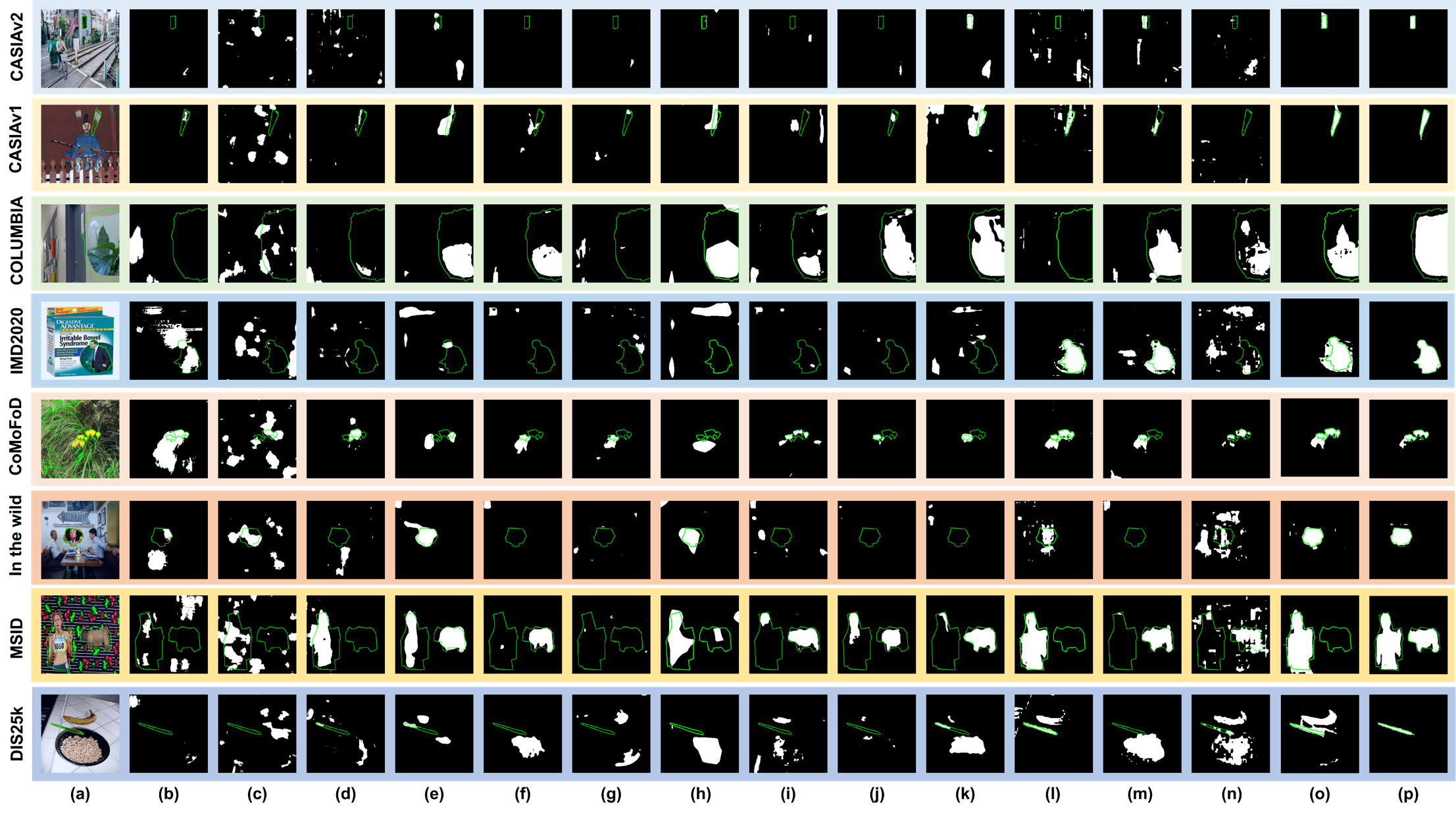}
  \caption{Qualitative comparison of other methods and SARIF under the CASIAv2 training scheme. (a) Input images with ground truth. (b) UNet \cite{ronneberger2015u}. (c) MantraNet \cite{wu2019mantra}. (d) RRUNet \cite{bi2019rru}. (e) TransForensic \cite{hao2021transforensics}. (f) FBINet \cite{gu2022fbi}. (g) MT-SENet \cite{zhang2021multi}. (h) MVSSNet \cite{dong2022mvss}. (i) PIMNet \cite{bai2025pim}. (j) EITLNet \cite{guo2024effective}. (k) M2SFormer \cite{nam2025m2sformer}. (l) SAM \cite{kirillov2023segment}. (m) autoSAM \cite{shaharabany2023autosam}. (n) IMDprompter \cite{zhang2025imdprompter}. (o) SAFIRE \cite{kwon2025safire}. (p) \textbf{SARIF(ours)}. \textcolor{green}{\textbf{Green}} lines denote the boundaries of the ground truth.}
  \label{fig:Qualitative_result}
\end{figure*}

\subsection{Comparison With State-of-the-art Method}
We evaluate our method against a diverse set of representative forgery localization baselines that span three backbone families: (i) CNN-based architectures such as UNet \cite{ronneberger2015u}, MantraNet \cite{wu2019mantra}, RRUNet \cite{bi2019rru}, TransForensic \cite{hao2021transforensics}, MT-SENet \cite{zhang2021multi}, MVSSNet \cite{dong2022mvss} and FBI-Net \cite{gu2022fbi} which rely on convolutional feature extractors; (ii) Transformer-based architectures such as EITLNet \cite{guo2024effective}, M2SFormer \cite{nam2025m2sformer}, PIMNet \cite{bai2025pim} which leverage global self-attention to capture long-range manipulation cues beyond local texture inconsistencies; and (iii) SAM-based architectures such as autoSAM \cite{shaharabany2023autosam}, IMDPROMPTER \cite{zhang2025imdprompter}, SAFIRE \cite{kwon2025safire} and the original SAM \cite{kirillov2023segment}, which adapt the Segment Anything Model for manipulation mask prediction. By comparing against all three categories, we ensure that our evaluation covers both traditional CNN-style forensic detectors, modern vision transformer models, and recent SAM-based models, demonstrating that our approach remains competitive across fundamentally different backbone designs.

\noindent \textbf{Quantitative Result.}
Tabs. \ref{tab:placeholder_dsc}-\ref{tab:placeholder_miou} report the cross-domain localization results. SARIF achieves the best seen-domain performance and strong average generalization across unseen datasets. On DSC, SARIF performs best on several unseen benchmarks, while SAFIRE remains stronger on DIS25K, IMD2020, and In the Wild, and M2SFormer is strongest on MISD. On mIoU, SARIF performs best on most datasets, whereas SAFIRE remains stronger on IMD2020 and In the Wild. These results indicate that SARIF is highly competitive across domains and achieves the strongest overall average performance.

\noindent \textbf{Qualitative Result.}
Fig. \ref{fig:Qualitative_result} presents qualitative comparisons under the CASIAv2-only training scheme. Compared with CNN- and transformer-based baselines, SARIF more often suppresses background leakage and produces sharper boundaries. Overall, SARIF tends to produce sharper and more spatially coherent forgery masks than the compared methods.
\begin{table}[t]
    \centering
    \caption{\textbf{Ablation study on Adapter, FSIE and FGMD.} $(\cdot)$ denotes the standard deviations of five-fold cross-validation experiment results.}
    \label{tab:setting_ablation}

    \resizebox{\linewidth}{!}{%
    \begin{tabular}{|c|c|c|c|c|c|ccc|ccc|}
        \hline
        \multirow{2}{*}{Setting} &
        \multirow{2}{*}{Adapter} &
        \multirow{2}{*}{FSIE} &
        \multirow{2}{*}{FGMD} &
        \multirow{2}{*}{Params(M)} &
        \multirow{2}{*}{FLOPs(G)} &
        \multicolumn{3}{c|}{Seen} &
        \multicolumn{3}{c|}{Unseen} \\ \cline{7-12}
        & & & & & &
        DSC & mIoU & AUC &
        DSC & mIoU & AUC \\ \hline

        1 &  &  &  & 13.90 & 490.65 &
        23.6 (9.8) & 18.1 (8.4) & 63.0 (4.6) &
        23.3 (6.6) & 16.7 (5.0) & 59.9 (3.5) \\

        2 & $\checkmark$ &  &  & 34.90 & 503.79 &
        41.2 (14.5) & 35.0 (13.6) & 71.4 (6.9) &
        33.0 (11.0) & 26.0 (9.7) & 64.6 (6.1) \\

        3 & $\checkmark$ & $\checkmark$ &  & 41.04 & 994.88 &
        62.4 (12.9) & 56.0 (13.3) & 83.0 (5.4) &
        44.7 (15.2) & 38.1 (14.8) & 70.2 (8.6) \\

        4 & $\checkmark$ &  & $\checkmark$ & 41.34 & 506.13 &
        62.0 (12.1) & 55.5 (12.4) & 82.5 (5.1) &
        42.6 (15.2) & 35.4 (14.5) & 69.1 (8.6) \\

        5 & $\checkmark$ & $\checkmark$ & $\checkmark$ & 52.62 & 998.43 &
        \textbf{\textcolor{red}{63.1 (12.9)}} &
        \textbf{\textcolor{red}{56.7 (13.3)}} &
        \textbf{\textcolor{red}{83.3 (5.4)}} &
        \textbf{\textcolor{red}{57.9 (7.1)}} &
        \textbf{\textcolor{red}{49.8 (6.1)}} &
        \textbf{\textcolor{red}{78.3 (3.6)}} \\
        \hline
    \end{tabular}%
    }
\end{table}
\subsection{Ablation Studies on FSIE and FGMD}

Under the CASIAv2 training protocol, we conduct ablation studies to isolate the contributions of the dual-encoder Forgery-Specific Information Extractor (FSIE) and the Feedback-Guided Mask Decoder (FGMD). Unless otherwise noted, all ablation settings use the same data split, preprocessing, training schedule, and evaluation protocol as the main experiments.

Tab.~\ref{tab:setting_ablation} summarizes the results. Starting from vanilla SAM, adding adapters already provides a substantial improvement, showing that lightweight adaptation is effective for forgery localization. Building on this adapter-based baseline, FSIE and FGMD each provide further gains from different perspectives: FSIE improves the quality of forgery-aware features delivered to the decoder, whereas FGMD improves the quality of the predicted masks through iterative refinement. Combining both modules yields the best overall performance, suggesting that forgery-specific cue extraction and feedback-based mask refinement are complementary.

\noindent\textbf{Setting 1 (Vanilla SAM).}
Directly applying SAM to forgery localization results in poor performance. Because of the domain gap between natural-object segmentation and image forensics, SAM often fails to attend to manipulation-specific traces, leading to inaccurate localization and noisy segmentation masks.

\noindent\textbf{Setting 2 (SAM + Adapters).}
Adding lightweight adapters to the SAM image encoder significantly improves performance over the vanilla SAM baseline. This result indicates that parameter-efficient adaptation helps the model encode forgery-relevant patterns and partially bridge the domain gap.

\noindent\textbf{Setting 3 (SAM + Adapters + FSIE).}
When FSIE is added on top of the adapter-based SAM, performance improves further. FSIE extracts forgery-sensitive residual cues from the dual encoder, organizes them into stage-wise prompts, and delivers them to the decoder. Exposing these multi-level cues helps the decoder better localize subtle manipulation artifacts.

\noindent\textbf{Setting 4 (SAM + Adapters + FGMD).}
Adding FGMD to the adapter-based SAM also improves performance. By feeding the predicted mask back as a mask prompt, FGMD progressively refines the output and reduces boundary ambiguity, resulting in cleaner masks. However, since FGMD mainly improves mask quality given the available cues, its gain is naturally limited when the underlying forgery-specific features are weak.

\noindent\textbf{Setting 5 (SARIF).}
The full model achieves the best overall performance. FSIE supplies hierarchical forgery-aware cues from the dual encoder, while FGMD refines the prediction through feedback-guided decoding. Their combination leads to more robust localization and cleaner segmentation, confirming that the two components are complementary.

\begin{figure}[t]
  \centering
  \begin{minipage}[t]{0.58\linewidth}
    \centering
    \includegraphics[width=\linewidth]{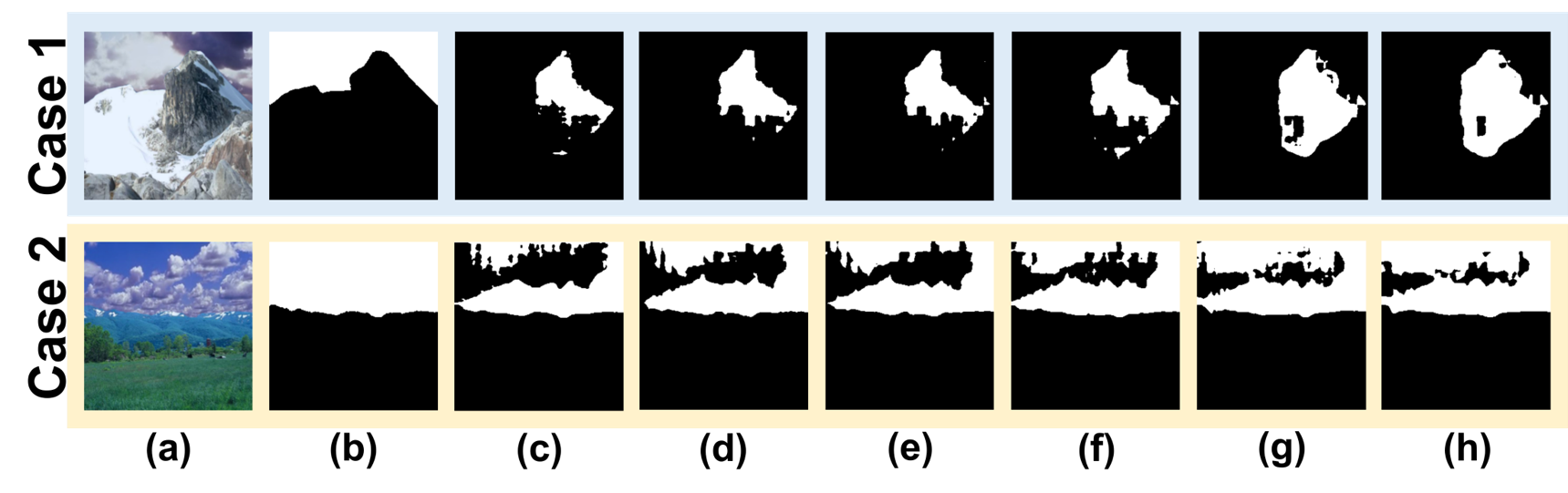}
    \captionof{figure}{\textbf{Failure cases}. (a) Input image. (b) Ground truth. (c) initial mask. (d$-$h) refinement stages 1 to 5. (h) indicates final prediction mask.}
    \label{fig:Fail_case}
  \end{minipage}\hfill
  \begin{minipage}[t]{0.41\linewidth}
    \centering
    \includegraphics[width=\linewidth]{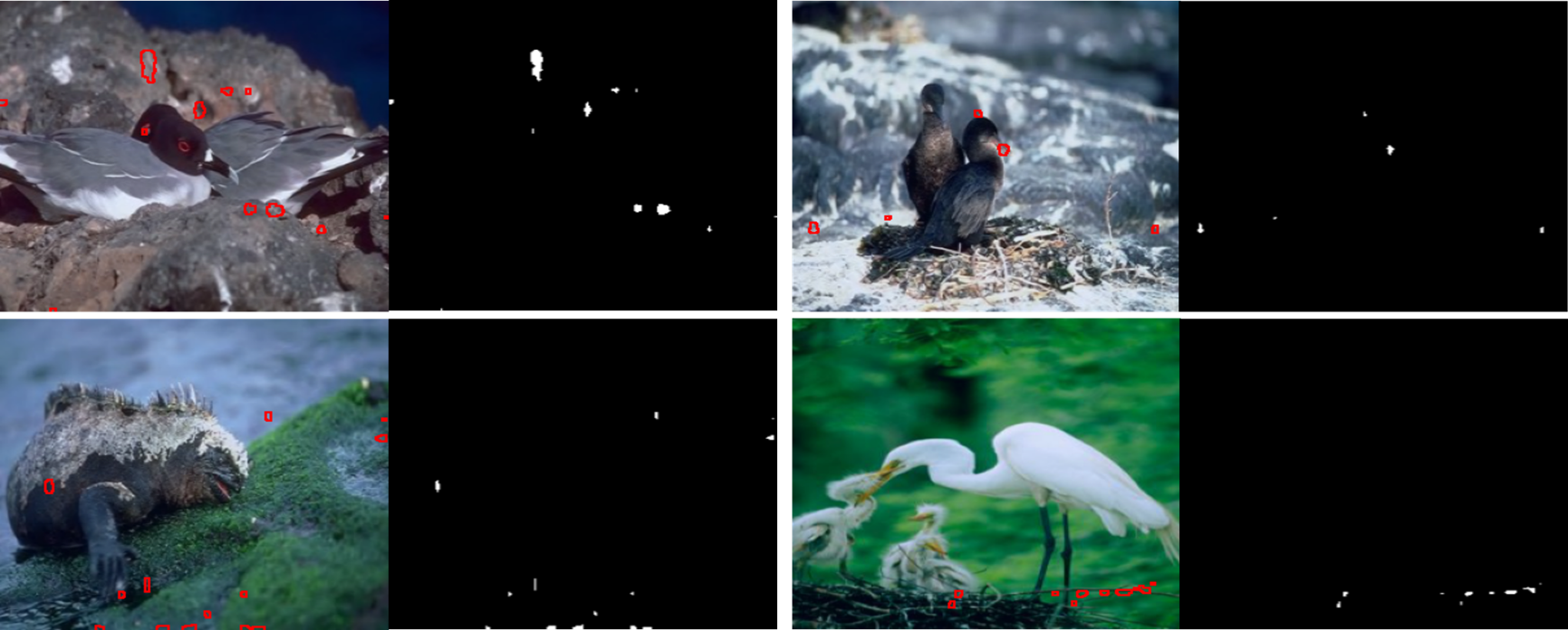}
    \captionof{figure}{\textbf{Authentic Image Prediction}. Left: Input image, Right: Model prediction.}
    \label{fig:Authentic}
  \end{minipage}
\end{figure}
\section{Discussion}

In this section, we focus on various questions that naturally arise when assessing SARIF’s practical usability.

\noindent \textit{\textbf{Q1. How does SARIF behave when provided pristine images?}}
Analyzing the behavior on pristine images is critical for practical forensic deployment. As shown in Fig. \ref{fig:Authentic}, when a fully authentic image is provided, SARIF doesn't predict an all-zero mask; instead, it may produce sparse low-level responses. Importantly, these responses tend to be \textit{\textbf{spatially scattered}}, which contrasts with the \textit{\textbf{spatially coherent masks}} typically observed on manipulated regions. This observation indicates that SARIF may produce sparse responses on pristine images; we therefore report pristine-image false-positive statistics in Appendix Section~\ref{appendix_False_Positive} to quantify this behavior.

\noindent \textit{\textbf{Q2. What are the limitations and typical failure cases of SARIF?}}
Fig. \ref{fig:Fail_case} presents failure cases of SARIF. \textbf{Case 1}. When the initial mask localizes an incorrect region, subsequent refinement tends to reinforce the error and continue refining the wrong area. This indicates that the adapted encoder may capture unreliable forgery cues, causing refinement to amplify an erroneous initial mask. As a result, overall performance largely depends on how effectively the adapter captures forgery cues. \textbf{Case 2}. The SAM-based architecture can introduce blocky artifacts because the prompt encoder first projects the mask prediction into a low-resolution embedding space, after which the mask decoder upsamples it with transposed convolution. This design can produce staircase-like artifacts.
\iffalse
\begin{figure}[t]
  \centering
  \begin{minipage}[t]{0.58\linewidth}
    \centering
    \includegraphics[width=\linewidth]{figure/Failure.png}
    \captionof{figure}{\textbf{Failure cases}. (a) Input image. (b) Ground truth. (c) initial mask. (d$-$h) iteration 1 to 5. (h) indicates final prediction mask.}
    \label{fig:Fail_case}
  \end{minipage}\hfill
  \begin{minipage}[t]{0.41\linewidth}
    \centering
    \includegraphics[width=\linewidth]{figure/Authentic_Image_Prediction.png}
    \captionof{figure}{\textbf{Authentic Image Prediction}. Left: Input image, Right: Model prediction.}
    \label{fig:Authentic}
  \end{minipage}
\end{figure}
\fi
\begin{figure}[t]
  \centering
  \includegraphics[width=\linewidth,height=0.3\textheight]{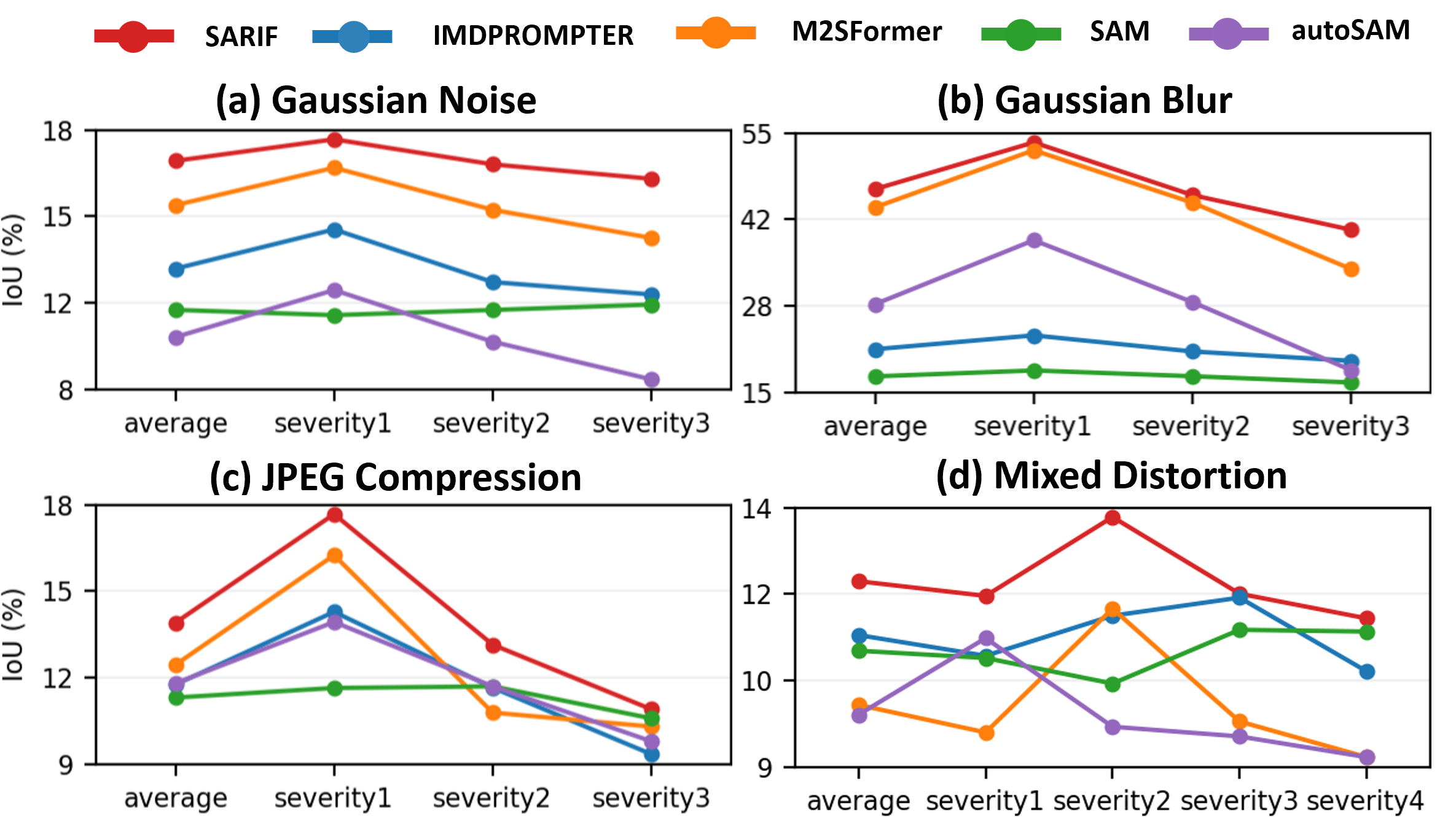}
  \caption{\textbf{Various distortion tests on CASIAv2}. (a) Gaussian Noise, severity 1, 2, 3 means $\sigma = $ 0.1, 0.3, 0.5. (b) Gaussian Blur, severity 1, 2, 3 means $\sigma =$ 3,5,9. (c) JPEG Compression, severity 1, 2, 3 means $q =$ 100, 50, 10. (d) Mixed robustness test, severity 1-4 correspond to (blur \& JPEG), (noise \& blur), (noise \& JPEG), (noise \& blur \& JPEG), respectively. For each mixed setting, severity level 2 was used for every individual distortion component.}
  \label{fig:distortion}
\end{figure}

\noindent \textit{\textbf{Q3. Can SARIF be reliably applied in real-world scenarios?}}
In real-world scenarios, images often suffer from hybrid and variable distortions \cite{wang2025diefgn}. To assess practical robustness, we evaluate SARIF under Gaussian noise, Gaussian blur, JPEG compression, and mixed distortions. As shown in Fig.~\ref{fig:distortion}, SARIF remains competitive across these settings and maintains strong performance under mixed distortions. These results suggest promising robustness in degraded environments.

\noindent \textit{\textbf{Q4. What is SARIF’s computational complexity, and how does it perform on recent challenging forgery datasets?}} We further evaluated SARIF on more recent and challenging forgery benchmarks, including an inpainting dataset (CocoGlide \cite{Guillaro_2023_CocoGlide}) and AI-generated forgery dataset (TGIF \cite{Mareen_2024_TGIF}) in Tab. \ref{tab:Extra_evaluation}. SARIF remains competitive on recent and challenging forgery benchmarks, including CocoGlide \cite{Guillaro_2023_CocoGlide} and TGIF, indicating strong reliability and generalization to modern forgery patterns. In addition, we report the overall model complexity in Table~\ref{tab:Extra_evaluation} and the complexity of each module in Table~\ref{tab:setting_ablation}. Although SARIF is computationally more expensive than the recent transformer-based M2SFormer, it consistently achieves clear accuracy gains across multiple datasets. We therefore explicitly present the accuracy--efficiency trade-off.

% \noindent \textit{\textbf{Q5. Is this experimental setting Fairness?}} Our evaluation approach—using identical hyperparameters, and data splits—strictly follows standard benchmarking practice, ensuring fairness by eliminating confounding factors from per-method tuning.
\begin{table}[t]
\centering
\caption{Model complexity and evaluation on recent forgery datasets}
\small
\setlength{\tabcolsep}{3pt}
\renewcommand{\arraystretch}{1.15}
\selectfont
\resizebox{\columnwidth}{!}{%
\begin{tabular}{|c|c|c|c|ccc|ccc|}
\hline
\multirow{2}{*}{Model} &
\multicolumn{3}{c|}{Model Complexity} &
\multicolumn{3}{c|}{CocoGlide \cite{Guillaro_2023_CocoGlide}} &
\multicolumn{3}{c|}{TGIF \cite{Mareen_2024_TGIF}} \\
\cline{2-10}
& Params(M) & FLOPs(G) & Inference(ms) & DSC & mIoU & AUC & DSC & mIoU & AUC \\
\hline
SAM \cite{kirillov2023segment}       & 13.90 & 490.65 & 18.8 & 17.4 (2.1) & 11.7 (1.6) & 54.9 (0.9) & 13.4 (1.6) & 9.1 (1.2) & 58.8 (1.2) \\
\hline
autoSAM \cite{shaharabany2023autosam}      & 158.53 & 1087.99 & 21.4 & 21.8 (1.0) & 14.9 (0.5) & 56.8 (0.3) & 16.9 (0.9) & 11.9 (0.8) & 61.2 (0.9) \\
\hline
IMDPROMPTER \cite{zhang2025imdprompter}  & 605.07  & 989.11  & 22.5 & 26.0 (1.4) & 17.9 (1.4) & 58.8 (0.7) & 18.6 (1.5) & 12.8 (1.3) & 61.6 (1.1) \\
\hline
M2SFormer \cite{nam2025m2sformer}    & 107.38   & 15.16  & 2.5     & 12.2 (1.6) & 7.7 (1.1)  & 52.0 (0.6) & 14.6 (0.8) & 10.3 (0.7) & 59.8 (0.3) \\
\hline
SARIF (Ours)   & 52.62 & 998.43  & 47.5 & \textbf{\textcolor{red}{34.0 (3.7)}} & \textbf{\textcolor{red}{27.3 (3.1)}} & \textbf{\textcolor{red}{63.7 (1.8)}} & \textbf{\textcolor{red}{30.8 (2.6)}} & \textbf{\textcolor{red}{23.7 (2.1)}} & \textbf{\textcolor{red}{68.7 (1.7)}} \\
\hline
\end{tabular}%
}
\label{tab:Extra_evaluation}
\end{table}

\section{Conclusion}
In this paper, we attempted to overcome the limitations of existing forgery localization methods, which are generalization performance and robustness by introducing SARIF—a fully automatic forgery localization framework that leverages SAM foundation-model for broad generalization and channels it into manipulation-aware prompting and refinement. The pipeline initializes from a prompt-free base mask produced by the mask decoder conditioned only on the fine-tuned SAM embeddings. Forgery-specific features are then derived by integrating the fine-tuned and original image embeddings, and combined with a mask prompt encoded from the previous prediction to drive the lightweight SAM mask decoder. Iterative, feedback-guided refinement with deep supervision at every stage improves boundary fidelity and stabilizes training, yielding consistent gains without manual points, boxes or masks. Experiments across heterogeneous datasets indicate strong cross-domain robustness, highlighting the benefit of forgery-specific prompting and feedback-guided mask decoder. Ultimately, by converting task-specific signal into automatic prompts and driving a feedback-guided refinement loop on SAM’s lightweight decoder, our method preserves SAM’s structural advantages while amplifying manipulation cues and cross-domain generalization and establishing a practical foundation for reliable, large-scale automated forgery localization.
\section*{Acknowledgements}

This work was supported in part by the National Research Foundation of Korea(NRF) grant funded by the Korea government(MSIT) (RS-2025-16065822) and in part by Inha University Research Grant.
\bibliographystyle{splncs04}
\bibliography{main}
\clearpage
\setcounter{page}{1}
\maketitlesupplementary

\begin{table}[]
    \centering
    \scriptsize
    \begin{tabular}{c|c||c|c}
    \hline
    Dataset Name & Total Images & Copy-Move & Splicing \\
    \hline
    CASIAv2 \cite{pham2019hybrid} & 5,123 & 3,274 & 1,849 \\
    CASIAv1 \cite{Dong2013}       & 920 & 459 & 461 \\
    DIS25k \cite{tahir2024deep}   & 24,964 & 0 & 24,964 \\
    Columbia \cite{hsu06crfcheck} & 180 & 0 & 180 \\
    IMD2020 \cite{Novozamsky_2020_WACV} & 2,010 & - & - \\ 
    CoMoFoD \cite{tralic2013comofod} & 260 & 260 & 0 \\
    In the Wild \cite{huh2018fighting} & 201 & 0 & 201 \\
    MISD \cite{kadam2021multiple} & 300 & 0 & 300 \\
    \hline
    \end{tabular}
    \caption{Summary of the datasets used in this paper.}
    \label{tab:dataset_summary}
\end{table}

\section{Dataset Descriptions}
\label{appendix_dataset_descriptions}

\begin{itemize}
    \item \textbf{CASIAv1} \cite{Dong2013} and \textbf{CASIAv2} \cite{pham2019hybrid}: The CASIAv1 dataset consists of JPG images with a resolution of 384 × 256, including 459 copy move and 461 splicing images. CASIAv2 is more complex than CASIAv1, containing 5,123 tampered ones which consists of 3274 copy move images and 1849 splicing images. The image sizes range from 320 × 240 to 800 × 600 and are available in multiple formats, such as uncompressed BMP and TIFF.

    \item \textbf{DIS25k} \cite{tahir2024deep}: The DIS25k dataset is a large-scale image splicing dataset designed to enhance the realism and complexity of manipulated images. It contains 24,964 spliced images, generated using image composition techniques, such as deep image matting and harmonization, to improve the seamlessness of manipulation. The dataset was created by leveraging the OPA dataset for rational object placement and refining the images with advanced blending techniques, making the forgeries harder to detect. The image sizes vary but generally range from 512 × 512 to 1920 × 1080, ensuring diversity in resolution.

    \item \textbf{Columbia} \cite{hsu06crfcheck}: The Columbia dataset is predominantly altered using splicing and contains high-resolution images. The manipulated regions often span large portions of scenes. Image sizes range from 757 × 568 to 1152 × 768 and are provided in TIFF or BMP formats. In total, it includes 180 tampered images.

    \item \textbf{IMD2020} \cite{Novozamsky_2020_WACV}: The IMD2020 dataset features a diverse set of tampered images collected from real-world sources on the internet, totaling approximately 2,010 manipulated samples. The images are provided in JPG and TIFF formats.

    \item \textbf{CoMoFoD} \cite{tralic2013comofod}: The CoMoFoD dataset is specifically designed for Copy-Move Forgery Detection (CMFD) and comprises 260 forged image sets, divided into two resolution categories: small (512 × 512) and large (3000 × 2000). The images are classified into five groups based on the type of manipulation applied: translation, rotation, scaling, combination, and distortion. Various post-processing techniques, including JPEG compression, blurring, noise addition, and color reduction, are applied to both the tampered and original images.

    \item  \textbf{In the Wild} \cite{huh2018fighting}: The In-the-Wild dataset is a collection of 201 manipulated images gathered from online sources such as THE ONION (a parody news website) and REDDIT PHOTOSHOP BATTLES (an online community focused on image manipulations). These images represent real-world, naturally occurring spliced forgeries. The images in this dataset come in various sizes, reflecting the diversity of online manipulations.

    \item \textbf{MISD} \cite{kadam2021multiple}: The Multiple Image Splicing Dataset (MSID) is the first publicly available dataset specifically designed for multiple image splicing detection. It contains 300 multiple spliced images, all in JPG format with a resolution of 384 × 256. The dataset was created by combining images from the CASIAv1 dataset and applying multiple splicing operations using Figma software. The spliced images feature various post-processing techniques such as rotation and scaling to enhance realism.
\end{itemize}

\section{Metrics Descriptions}
\label{appendix_metric_descriptions}

In this section, we describe the metrics used in this paper. For convenience, we denote $TP, FP$, and $FN$ as the number of samples of true positive, false positive, and false negative between two binary masks $A$ and $B$.  

\begin{itemize}
    \item The \textit{Mean Dice Similarity Coefficient (DSC)} \cite{milletari2016v} measures the similarity between two samples and is widely used to assess the performance of segmentation tasks, such as image segmentation or object detection. \textbf{\underline{Higher is better}}. For given two binary masks $A$ and $B$, DSC is defined as follows:
    \begin{equation}
        \textbf{DSC}(A, B) = \frac{2 \times | A \cap B |}{| A \cup B |} = \frac{2 \times TP}{2 \times TP + FP + FN}
    \end{equation}

    \item The \textit{Mean Intersection over Union (mIoU)} measures the ratio of the intersection area to the union area between predicted and ground truth masks in segmentation tasks. \textbf{\underline{Higher is better}}. For given two binary masks $A$ and $B$, mIoU is defined as follows:
    \begin{equation}
        \textbf{mIoU}(A, B) = \frac{A \cap B}{A \cup B} = \frac{TP}{TP + FP + FN}
    \end{equation}
    
    \item The \textit{Area Under the Curve (AUC)} measures the overall discriminative ability of a model across all possible decision thresholds and is particularly useful for evaluating binary classification or pixel-wise detection tasks where threshold selection can bias results. \textbf{\underline{Higher is better}}. AUC summarizes the relationship between the True Positive Rate (TPR) and False Positive Rate (FPR) by computing the area under the Receiver Operating Characteristic (ROC) curve. Formally, it is expressed as:
    \begin{equation}
    \textbf{AUC} = \int_{0}^{1} \text{TPR}(t), d(\text{FPR}(t))
    \end{equation}
    where $\text{TPR}(t) = \frac{TP}{TP + FN}$ and $\text{FPR}(t) = \frac{FP}{FP + TN}$ at threshold $t$. A higher AUC value indicates that the model achieves better separation between the manipulated (positive) and authentic (negative) samples across varying decision thresholds.
\end{itemize}

\section{False positives on pristine images.}
\label{appendix_False_Positive}

\begin{table}[t]
    \centering
    \caption{Segmentation results on CASIAv2 \cite{pham2019hybrid}. mIoU is reported on the seen-domain CASIAv2, while FPR is reported on CASIAv2 authentic \cite{pham2019hybrid}. $(\cdot)$ denotes the standard deviations of five-fold cross-validation experiment results. \textcolor{red}{\textbf{{Red}}} and \textcolor{blue}{\textbf{{Blue}}} denote the best and second best results.}
    \tiny
    \renewcommand{\arraystretch}{1.25}
    \begin{tabular}{c|c|c|c}
        \hline
        \cellcolor[gray]{.9} &
        \cellcolor[gray]{.9} &
        \multicolumn{1}{c|}{\cellcolor[gray]{.9}Seen Domain} &
        \multicolumn{1}{c}{\cellcolor[gray]{.9}Authentic} \\
        \cline{3-4}

        \cellcolor[gray]{.9} &
        \cellcolor[gray]{.9} &
        \multicolumn{1}{c|}{\cellcolor[gray]{.9}CASIAv2 \cite{pham2019hybrid}} &
        \multicolumn{1}{c}{\cellcolor[gray]{.9}CASIAv2\_Authentic \cite{pham2019hybrid}} \\
        \cline{3-4}

        \multirow{-3}{*}{\cellcolor[gray]{.9}Type} &
        \multirow{-3}{*}{\cellcolor[gray]{.9}Method} &
        \cellcolor[gray]{.9}mIoU &
        \cellcolor[gray]{.9}FPR \\
        \hline

        \rowcolor{red!10}
        & UNet \cite{ronneberger2015u}
        & 25.8 (9.2)
        & 22.4 (5.9) \\

        \rowcolor{red!10}
        & MantraNet \cite{wu2019mantra}
        & 11.9 (5.7)
        & 9.6 (2.9) \\

        \rowcolor{red!10}
        & RRUNet \cite{bi2019rru}
        & 15.8 (8.4)
        & 10.8 (5.6) \\

        \rowcolor{red!10}
        & TransForensic \cite{hao2021transforensics}
        & 32.0 (13.9)
        & 17.0 (7.3) \\

        \rowcolor{red!10}
        & FBINet \cite{gu2022fbi}
        & 29.2 (12.8)
        & \textbf{\textcolor{red}{3.0 (0.7)}} \\

        \rowcolor{red!10}
        & MT-SENet \cite{zhang2021multi}
        & 14.6 (7.6)
        & 5.4 (0.4) \\

        \rowcolor{red!10}
        \multirow{-7}{*}{\rotatebox[origin=c]{90}{\textbf{Convolution}}}
        & MVSSNet \cite{dong2022mvss}
        & 23.4 (11.1)
        & 10.2 (4.7) \\
        \hline

        \rowcolor{orange!10}
        & PIMNet \cite{bai2025pim}
        & 48.5 (14.6)
        & 15.5 (4.7) \\

        \rowcolor{orange!10}
        & EITLNet \cite{guo2024effective}
        & 47.9 (14.7)
        & 5.5 (0.3) \\

        \rowcolor{orange!10}
        \multirow{-3}{*}{\rotatebox[origin=c]{90}{\textbf{Trans.}}}
        & M2SFormer \cite{nam2025m2sformer}
        & \textcolor{blue}{\textbf{\textit{50.8 (12.8)}}}
        & 10.1 (1.2) \\
        \hline

        \rowcolor{green!10}
        & SAM \cite{kirillov2023segment}
        & 21.0 (10.5)
        & 10.5 (0.8) \\

        \rowcolor{green!10}
        & AutoSAM \cite{shaharabany2023autosam}
        & 41.7 (19.0)
        & 8.5 (1.5) \\

        \rowcolor{green!10}
        & IMDPromter \cite{zhang2025imdprompter}
        & 25.9 (14.9)
        & 9.4 (1.3) \\

        \rowcolor{green!10}
        & SAFIRE \cite{kwon2025safire}
        & 45.5 (10.9)
        & 11.6 (2.7) \\

        \rowcolor{green!10}
        \multirow{-5}{*}{\rotatebox[origin=c]{90}{\textbf{SAM}}}
        & \textbf{SARIF (Ours)}
        & \textbf{{\textcolor{red}{56.7 (13.3)}}}
        & \textcolor{blue}{\textbf{5.1 (0.8)}} \\
        \hline
    \end{tabular}
    \vspace{-0.5em}
    \label{tab:casiav2_miou_fpr}
\end{table}
For pristine images, the ground-truth mask is entirely zero; therefore, any activated pixel is counted as a false positive. In Table~\ref{tab:casiav2_miou_fpr}, the FPR on the CASIAv2 authentic set \cite{pham2019hybrid} can therefore be interpreted as a compact measure of the residual response each method produces on authentic inputs. Read together with Fig.~\ref{fig:Authentic}, this metric complements the qualitative observation that SARIF may still yield weak activations on pristine images, while these responses remain limited rather than forming broad, spatially coherent masks. The same table also reports mIoU on the seen-domain, allowing localization accuracy on manipulated images and false responses on authentic images to be considered jointly.

Under the training scheme based on CASIAv2 \cite{pham2019hybrid}, SARIF achieves the highest mIoU while also attaining the lowest FPR among the SAM-based methods and the second-lowest overall on the authentic set. Among the methods with similarly low false-positive behavior, SARIF still provides the strongest localization performance, outperforming EITLNet \cite{guo2024effective}, FBINet \cite{gu2022fbi}, and MT-SENet \cite{zhang2021multi}. Compared with FBINet \cite{gu2022fbi}, SARIF improves mIoU by 27.5 points with only a 2.1 percentage-point increase in FPR. Taken together, these results support the qualitative observation that SARIF maintains strong localization performance while keeping responses on pristine images relatively limited.

\section{Additional Threshold-Independent Evaluation}
\label{app:auc_analysis}

\begin{table}[t]
    \centering
    \caption{Representative AUC results under the CASIAv2 \cite{pham2019hybrid} training scheme. AUC is reported on the seen-domain CASIAv2 and seven unseen-domain datasets. $(\cdot)$ denotes the standard deviations of five-fold cross-validation experiment results. \textcolor{red}{\textbf{{Red}}} and \textcolor{blue}{\textbf{{Blue}}} denote the best and second best results.}
    \label{tab:casia_auc}
    \tiny
    \setlength{\tabcolsep}{3.0pt}
    \renewcommand{\arraystretch}{1.25}
    \resizebox{\linewidth}{!}{%
    \begin{tabular}{c|l|c|c|c|c|c|c|c|c}
        \hline
        \cellcolor[gray]{.9} &
        \cellcolor[gray]{.9} &
        \multicolumn{1}{c|}{\cellcolor[gray]{.9}Seen Domain} &
        \multicolumn{7}{c}{\cellcolor[gray]{.9}Unseen Domain} \\
        \cline{3-10}

        \cellcolor[gray]{.9} &
        \cellcolor[gray]{.9} &
        \multicolumn{1}{c|}{\cellcolor[gray]{.9}CASIAv2 \cite{pham2019hybrid}} &
        \multicolumn{1}{c|}{\cellcolor[gray]{.9}DIS25K \cite{tahir2024deep}} &
        \multicolumn{1}{c|}{\cellcolor[gray]{.9}CASIAv1 \cite{Dong2013}} &
        \multicolumn{1}{c|}{\cellcolor[gray]{.9}Columbia \cite{hsu06crfcheck}} &
        \multicolumn{1}{c|}{\cellcolor[gray]{.9}IMD2020 \cite{Novozamsky_2020_WACV}} &
        \multicolumn{1}{c|}{\cellcolor[gray]{.9}CoMoFoD \cite{tralic2013comofod}} &
        \multicolumn{1}{c|}{\cellcolor[gray]{.9}In the Wild \cite{huh2018fighting}} &
        \multicolumn{1}{c}{\cellcolor[gray]{.9}MSID \cite{kadam2021multiple}} \\
        \cline{3-10}

        \multirow{-3}{*}{\cellcolor[gray]{.9}\textbf{Type}} &
        \multirow{-3}{*}{\cellcolor[gray]{.9}\textbf{Method}} &
        \cellcolor[gray]{.9}AUC &
        \cellcolor[gray]{.9}AUC &
        \cellcolor[gray]{.9}AUC &
        \cellcolor[gray]{.9}AUC &
        \cellcolor[gray]{.9}AUC &
        \cellcolor[gray]{.9}AUC &
        \cellcolor[gray]{.9}AUC &
        \cellcolor[gray]{.9}AUC \\
        \hline

        \rowcolor{red!10}
        \rotatebox[origin=c]{90}{\textbf{Conv.}}
        & MVSSNet \cite{dong2022mvss}
        & 69.2 (8.1)
        & 64.9 (2.8)
        & 69.7 (1.6)
        & 58.8 (1.3)
        & 64.4 (2.2)
        & 59.9 (2.2)
        & 60.6 (1.9)
        & 73.6 (1.6) \\
        \hline

        \rowcolor{orange!10}
        \rotatebox[origin=c]{90}{\textbf{Trans.}}
        & M2SFormer \cite{nam2025m2sformer}
        & \textbf{\textcolor{red}{83.8 (5.1)}}
        & 71.7 (1.4)
        & \textbf{\textcolor{red}{81.3 (0.8)}}
        & 64.7 (3.4)
        & 68.5 (1.4)
        & 64.7 (1.2)
        & 64.7 (0.9)
        & \textbf{\textcolor{red}{79.5 (0.6)}} \\
        \hline

        \rowcolor{green!10}
        & IMDPrompter \cite{zhang2025imdprompter}
        & 66.8 (7.3)
        & 65.8 (1.3)
        & 72.0 (1.6)
        & 59.6 (3.1)
        & 64.4 (1.4)
        & 60.9 (0.8)
        & 63.0 (2.3)
        & 67.0 (1.4) \\

        \rowcolor{green!10}
        & SAFIRE \cite{kwon2025safire}
        & 79.7 (1.8)
        & \textcolor{blue}{\textbf{\textit{76.0 (0.8)}}}
        & 78.5 (0.4)
        & \textcolor{blue}{\textbf{\textit{69.7 (2.4)}}}
        & \textbf{\textcolor{red}{79.4 (0.7)}}
        & \textcolor{blue}{\textbf{\textit{73.5 (0.8)}}}
        & \textbf{\textcolor{red}{81.2 (1.3)}}
        & 67.9 (0.7) \\

        \rowcolor{green!10}
        \multirow{-3}{*}{\rotatebox[origin=c]{90}{\textbf{SAM}}}
        & \textbf{SARIF (Ours)}
        & \textcolor{blue}{\textbf{\textit{83.3 (5.4)}}}
        & \textbf{\textcolor{red}{76.1 (1.5)}}
        & \textcolor{blue}{\textbf{\textit{79.3 (0.9)}}}
        & \textbf{\textcolor{red}{73.2 (1.7)}}
        & \textcolor{blue}{\textbf{\textit{77.1 (1.2)}}}
        & \textbf{\textcolor{red}{83.8 (1.2)}}
        & \textcolor{blue}{\textbf{\textit{76.2 (2.3)}}}
        & \textcolor{blue}{\textbf{\textit{77.8 (1.0)}}} \\
        \hline
    \end{tabular}}
    \vspace{-0.5em}
\end{table}

Table~\ref{tab:casia_auc} reports representative AUC results under the CASIAv2 training scheme. 
AUC evaluates threshold-independent pixel-level separability and therefore complements the DSC and mIoU results reported in the main paper. 
SARIF achieves the best AUC on DIS25K, Columbia, and CoMoFoD, and obtains the second-best result on CASIAv2, CASIAv1, IMD2020, In the Wild, and MSID. 
These results indicate that SARIF maintains strong discriminative ability across both seen and unseen domains, rather than relying on a particular binarization threshold. 
Although SAFIRE shows stronger AUC on IMD2020 and In the Wild, and M2SFormer performs best on CASIAv2, CASIAv1, and MSID, SARIF remains consistently competitive across all evaluated datasets, supporting its cross-domain robustness.

\section{Refinement-Stage Analysis}
\label{app:refinement_stage}

\begin{table}[t]
    \centering
    \caption{IoU/DSC results across refinement stages. Intermediate masks are evaluated without retraining. IoU and DSC are reported in the form of IoU/DSC. \textcolor{red}{\textbf{{Red}}} denotes the best or tied-best result.}
    \label{tab:refinement_stage}
    \tiny
    \renewcommand{\arraystretch}{1.25}
    \begin{tabular}{c|c|c|c|c}
        \hline
        \cellcolor[gray]{.9} &
        \multicolumn{1}{c|}{\cellcolor[gray]{.9}Seen Domain} &
        \multicolumn{3}{c}{\cellcolor[gray]{.9}Unseen Domain} \\
        \cline{2-5}

        \cellcolor[gray]{.9} &
        \multicolumn{1}{c|}{\cellcolor[gray]{.9}CASIAv2 \cite{pham2019hybrid}} &
        \multicolumn{1}{c|}{\cellcolor[gray]{.9}DIS25K \cite{tahir2024deep}} &
        \multicolumn{1}{c|}{\cellcolor[gray]{.9}CASIAv1 \cite{Dong2013}} &
        \multicolumn{1}{c}{\cellcolor[gray]{.9}IMD2020 \cite{Novozamsky_2020_WACV}} \\
        \cline{2-5}

        \multirow{-3}{*}{\cellcolor[gray]{.9}\textbf{Stage}} &
        \cellcolor[gray]{.9}IoU/DSC &
        \cellcolor[gray]{.9}IoU/DSC &
        \cellcolor[gray]{.9}IoU/DSC &
        \cellcolor[gray]{.9}IoU/DSC \\
        \hline

        Init mask
        & 53.3/61.9
        & 39.6/45.8
        & 50.2/57.3
        & 38.2/46.6 \\

        5-block
        & 54.0/62.5
        & 40.3/46.5
        & 51.0/57.7
        & 38.7/47.5 \\

        11-block
        & 54.4/62.8
        & 41.0/47.1
        & 51.5/58.0
        & 39.3/47.9 \\

        17-block
        & 55.6/62.8
        & 41.3/47.4
        & 51.6/58.2
        & 39.6/48.0 \\

        23-block
        & 56.4/63.0
        & 41.5/47.7
        & 51.8/58.3
        & 40.0/48.2 \\

        Final mask
        & \textbf{\textcolor{red}{56.7/63.1}}
        & \textbf{\textcolor{red}{41.5/47.8}}
        & \textbf{\textcolor{red}{52.0/58.4}}
        & \textbf{\textcolor{red}{40.4/48.4}} \\
        \hline
    \end{tabular}
    \vspace{-0.5em}
\end{table}

Table~\ref{tab:refinement_stage} shows that the mask quality improves progressively across refinement stages. 
Starting from the initial mask, both IoU and DSC generally increase as the feedback-guided decoder receives additional forgery-specific cues from the selected SAM-ViT global-attention blocks. 
The final mask achieves the best or tied-best result on all evaluated datasets, indicating that the proposed refinement process accumulates useful spatial evidence rather than merely repeating the same decoding operation. 
The gains become smaller in later stages, suggesting a saturating refinement behavior and an accuracy--efficiency trade-off when earlier stopping is required.

\section{Technical Novelty and Positioning}
\label{app:technical_novelty}

To clarify the technical distinction of SARIF from recent SAM-based forgery localization methods, we provide a method-specific comparison with IMDPrompter, SAMIF, and SAFIRE. 
IMDPrompter learns multi-view prompts and selects effective prompts for image manipulation detection, whereas SARIF derives its prompts from the residual features between frozen and LoRA-adapted SAM encoders. 
SAMIF incorporates a high-frequency branch to adapt SAM for inpainting forensics, while SARIF focuses on extracting task-specific residual cues from a dual-encoder design. 
SAFIRE performs grid-prompted source-region segmentation and aggregates the resulting predictions, whereas SARIF refines masks through feedback-guided prompting using the previous prediction as a mask prompt. 
Therefore, SARIF is not merely another automatic prompt-generation variant; its core novelty lies in coupling dual-encoder residual extraction with feedback-guided prompting for robust image forgery localization.

\end{document}